\documentclass[10pt,twocolumn,letterpaper]{article}

\usepackage{iccv}
\usepackage{times}
\usepackage{epsfig}
\usepackage{graphicx}
\usepackage{amsmath}
\usepackage{amssymb}
\usepackage{comment}
\usepackage{booktabs}
\usepackage{multirow}
\usepackage{setspace}
\usepackage[breaklinks=true,bookmarks=false]{hyperref}

\iccvfinalcopy % *** Uncomment this line for the final submission

\def\iccvPaperID{465} % *** Enter the ICCV Paper ID here

\makeatletter
\DeclareRobustCommand\onedot{\futurelet\@let@token\@onedot}
\def\@onedot{\ifx\@let@token.\else.\null\fi\xspace}

\def\eg{\emph{e.g}\onedot}

\def\etc{\emph{etc}\onedot}

\makeatother

\begin{document}

\title{Webly Supervised Learning of Convolutional Networks}

\author{Xinlei Chen\\
Carnegie Mellon University\\
{\tt\small xinleic@cs.cmu.edu}
\and
Abhinav Gupta\\
Carnegie Mellon University\\
{\tt\small abhinavg@cs.cmu.edu}
}

% \setstretch{0.96}

\maketitle

\begin{abstract}
% In the last few years, we have made enormous progress in learning visual representations via convolutional neural networks (CNNs). We believe CNNs get their edge due to their ability to imbibe large amounts of data. Therefore, as we move forward a key question arises: how do we move from million image datasets to billion image counterparts? Do we continue to manually label images with the hope of scaling up the labeling to billion images? It is in this context that webly supervised learning assumes huge importance: if we can exploit the images on the web for training CNNs without manually labeling them, it will be a win-win for everyone. 
	We present an approach to utilize large amounts of web data for learning CNNs. Specifically inspired by curriculum learning, we present a two-step approach for CNN training. First, we use easy images to train an initial visual representation. We then use this initial CNN and adapt it to harder, more realistic images by leveraging the structure of data and categories. We demonstrate that our two-stage CNN outperforms a fine-tuned CNN trained on ImageNet on Pascal VOC 2012. %without even using a single ImageNet training label. 
We also demonstrate the strength of webly supervised learning by localizing objects in web images and training a R-CNN style~\cite{girshick2014rich} detector. It achieves the best performance on VOC 2007 where no VOC training data is used. Finally, we show our approach is quite robust to noise and performs comparably even when we use image search results from March 2013 (pre-CNN image search era).  
	
%Experiments on scene classification further justified the effectiveness of webly supervised CNNs.
\end{abstract}

\vspace{-0.1in}
\section{Introduction}
%With an enormous amount of visual data online, web and social media are among the most important sources of data for vision research. Vision datasets such as ImageNet~\cite{russakovsky2014imagenet}, PASCAL VOC~\cite{everingham2010pascal} and MS COCO~\cite{lin2014microsoft} have been created from Google or Flickr by harnessing human intelligence to filter out the noisy images returned by search engines. The resulting clean data has helped significantly advance performance on relevant tasks~\cite{lsvm-pami,krizhevsky2012imagenet,girshick2014rich,zhou2014learning}. For example, training a neural network on ImageNet followed by fine-tuning on PASCAL VOC has led to the state-of-the-art performance on the object detection challenge~\cite{krizhevsky2012imagenet,girshick2014rich}. But human supervision comes with a cost and its own problems (\eg inconsistency, incompleteness and bias~\cite{torralba2011unbiased}). Therefore, an alternative, and more appealing way is to learn visual representations from the web data directly, without using any manual labeling. But the big question is, can we actually use millions of images online without using any human supervision?

With an enormous amount of visual data online, web and social media are among the most important sources of data for vision research. Vision datasets such as ImageNet~\cite{russakovsky2014imagenet}, PASCAL VOC~\cite{everingham2010pascal} and MS COCO~\cite{lin2014microsoft} have been created from Google or Flickr by harnessing human intelligence to filter out the noisy images and label object locations. The resulting clean data has helped significantly advance performance on relevant tasks~\cite{lsvm-pami,krizhevsky2012imagenet,girshick2014rich,zhou2014learning}. For example, training a neural network on ImageNet followed by fine-tuning on PASCAL VOC has led to the state-of-the-art performance on the object detection challenge~\cite{krizhevsky2012imagenet,girshick2014rich}. But human supervision comes with a cost and its own problems (\eg inconsistency, incompleteness and bias~\cite{torralba2011unbiased}). Therefore, an alternative, and more appealing way is to learn visual representations and object detectors from the web data directly, without using any manual labeling of bounding boxes. But the big question is, can we actually use millions of images online without using any human supervision?

\begin{figure}[t]
\centering
\includegraphics[width=0.97\linewidth]{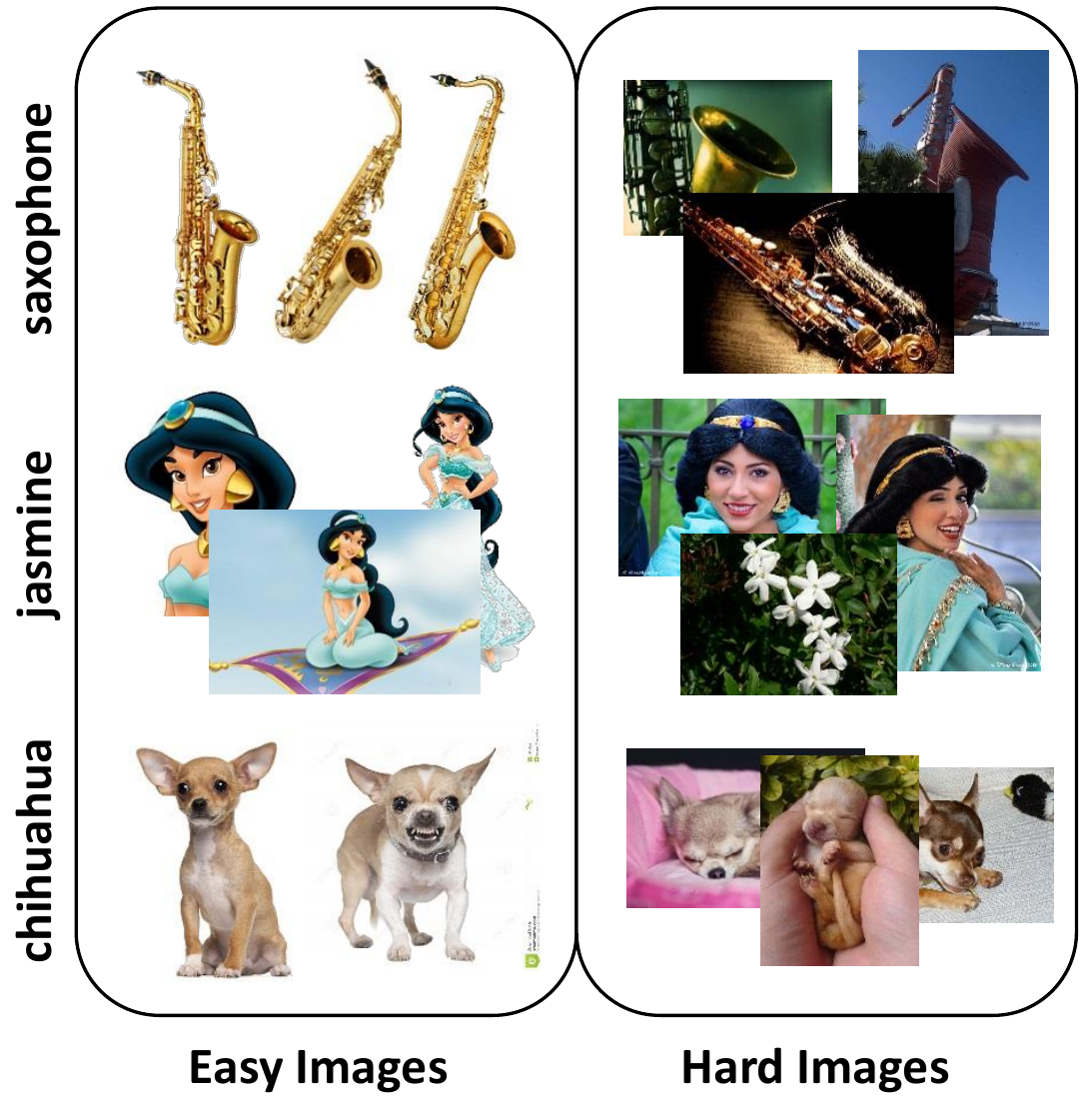}
\caption{{\small We investigate the problem of training a webly supervised CNN. Two types of visual data are available online: image search engine results (left) and photo-sharing websites (right). We train a two-stage network bootstrapping from clean examples retrieved by Google, and enhanced by noisier images from Flickr.}\label{fig:teaser}}
\vspace{-0.2in}
\end{figure}

In fact, researchers have pushed hard to realize this dream of learning visual representations and object detectors from web data. These efforts have looked at different aspects of webly supervised learning such as:
\begin{itemize}
\setlength\itemsep{0em}
\item {\bf What are the good sources of data?} Researchers have tried various search engines ranging from  text/image search engines~\cite{berg2006animals,wang2008annotating,vijayanarasimhan2008keywords,fergus2010learning} to Flickr images~\cite{ordonez2011im2text}.
\item {\bf What types of data can be exploited?} Researchers have tried to explore different types of data, like images-only~\cite{li2010optimol,chen2013neil}, images-with-text~\cite{berg2006animals,schroff2011harvesting} or even images-with-$n$-grams~\cite{divvala2014learning}).
\item {\bf How do we exploit the data?} Extensive algorithms (\eg probabilistic models~\cite{fergus2010learning,li2010optimol}, exemplar based models~\cite{chen2013neil}, deformable part models~\cite{divvala2014learning}, self organizing map~\cite{golge2014conceptmap} \etc) have been developed.
\item {\bf What should we learn from web data?} There has been lot of effort ranging from just cleaning data~\cite{fan2010harvesting,xia2014well,ordonez2011im2text} to training visual models~\cite{li2010optimol,torresani2010efficient,li2013harvesting}, to even discovering common-sense relationships~\cite{chen2013neil}.
\end{itemize}
Nevertheless, while many of these systems have seen orders of magnitudes larger number of images, their performance has never matched up against contemporary methods that receive extensive supervision from humans. Why is that?

Of course the biggest issue is the data itself: 1) it contains noise, and 2) is has bias - image search engines like Google usually operate in the high-precision low-recall regime and tend to be biased toward images where a single object is centered with a clean background and a canonical viewpoint~\cite{mezuman2012learning,berg2009finding,lin2014microsoft}. But is it just the data? We argue that it is not just the data itself, but also the ability of algorithms to learn from large data sources and generalize. For example, traditional approaches which use hand-crafted features (\eg HOG~\cite{chen2013neil}) and classifiers like support vector machines~\cite{divvala2014learning} have very few parameters (less capacity to memorize) and are therefore unlikely to effectively use large-scale training data. On the other hand, memory based nearest neighbors classifiers can better capture the distribution given a sufficient amount of data, but are less robust to the noise. Fortunately, Convolutional Neural Networks (CNNs)~\cite{krizhevsky2012imagenet} have resurfaced as a powerful tool for learning from large-scale data: when trained with ImageNet~\cite{russakovsky2014imagenet} ($\sim$1M images), it is not only able to achieve state-of-the-art performance for the same image classification task, but the learned representation can be readily applied to other relevant tasks~\cite{girshick2014rich,zhou2014learning}. % Delving inside the model, it was shown to actually learn concept detectors in its mid-level neurons that are relevant to the categories of interest~\cite{zeiler2014visualizing,zhou2014learning}. 

Attracted by their amazing capability to harness large-scale data, in this paper, we investigate webly supervised learning for CNNs (See Figure~\ref{fig:teaser}). Specifically, 1) we present a two-stage webly supervised approach to learning CNNs. First we show that CNNs can be readily trained for easy categories using images retrieved by search engines. We then adapt this network to hard (Flickr style) web images using the relationships discovered in easy images. 2) We show webly supervised CNNs also generalize well to relevant vision tasks, giving state-of-the-art performance compared to ImageNet pretrained CNNs if there is enough data. 3) We show state-of-the-art performance on VOC data for the scenario where not a single VOC training image is used - just the images from the web. 4) We also show competitive results on scene classification. We believe this paper opens up avenues for exploitation of web data to achieve next cycle of performance gain in vision tasks (and at no human labeling costs!).

\subsection{Why Webly Supervised?} 
Driven by CNNs, the field of object detection has seen a dramatic churning in the past two years, which has resulted in a significant improvement in the state-of-the-art performance. But as we move forward, how do we further improve performance of CNN-based approaches? We believe there are two directions. The first and already explored area is designing deeper networks~\cite{simonyan2014very,szegedy2014going}. We believe a more promising direction is to feed more data into these networks (in fact, deeper networks would often need more data to train). But more data needs more human labeling efforts. But data labeling in terms of bounding boxes can be very cumbersome and expensive. Therefore, if we can exploit web data for training CNNs, it would help us move from million to billion image datasets in the future. In this paper, we take the first step in demonstrating: 1) CNNs can be trained effectively by just exploiting web data at much larger scales; 2) competitive object detection results can be obtained without using a single bounding box labels from humans.

\begin{figure*}[t]
\centering
\includegraphics[width=1.0\linewidth]{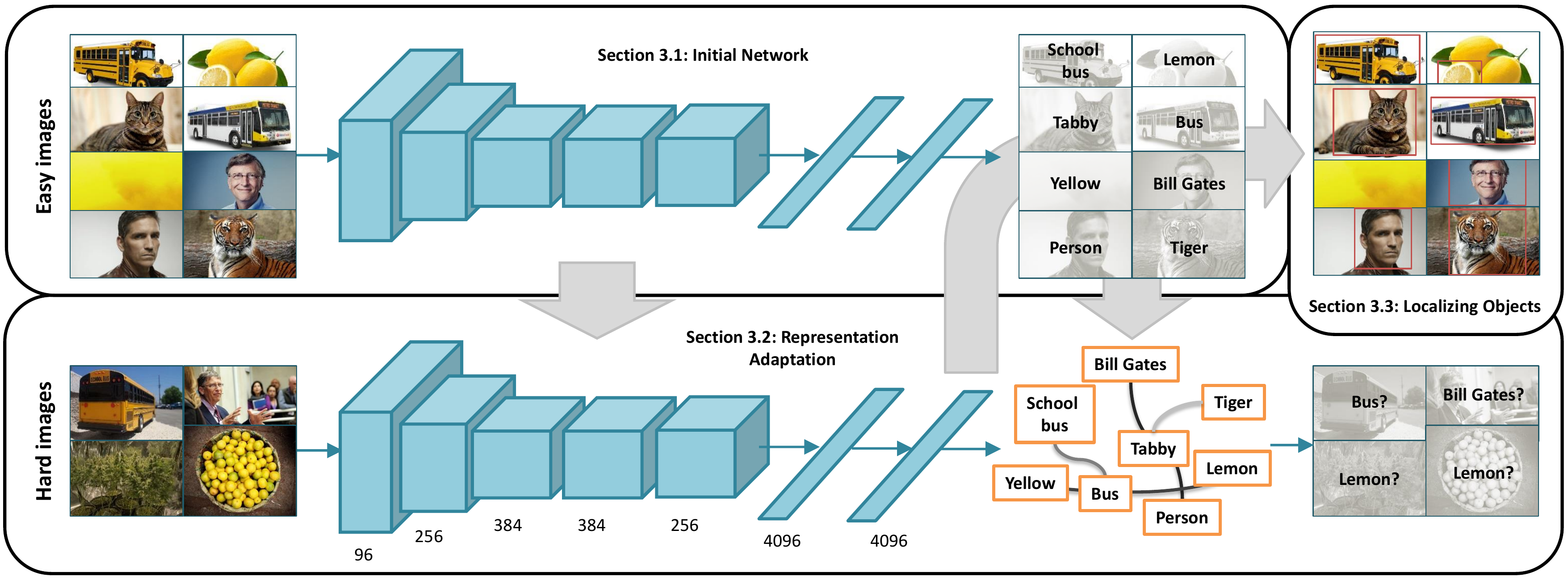}
\caption{{\small Outline of our approach. We first train a CNN using easy images from Google (above). This CNN is then used to find relationships and initialize another CNN (below) for harder images. The learned representations are in turn used to localize objects and clean up data.}\label{fig:outline}}
\vspace{-0.2in}
\end{figure*}

\vspace{-0.1in}
\section{Related Work}
Mining high-quality visual data and learning good visual representation for recognition from the web naturally form two aspects of a typical chicken-and-egg problem in vision. On one hand, clean and representative seed images can help build better and more powerful models; but on the other hand, models that recognize concepts well are crucial for indexing and retrieving image sets that contain the concept of interest. How to attack this problem has long been attractive to both industry and academia.

\noindent \textbf{From Models to Data}: Image retrieval~\cite{smeulders2000content,sivic2003video} is a classical problem in this setting. It is not only an active research topic, but also fascinating to commercial image search engines and photo-sharing websites since they would like to better capture data streams on the Internet and thus better serve user's information need. Over the years, various techniques (\eg click-through data) have been integrated to improve search engine results. Note that, using pretrained models (\eg CNN~\cite{xia2014well}) to clean up web data also falls into this category, since extensive human supervision has already been used.

\noindent \textbf{From Data to Models}: A more interesting and challenging direction is the opposite - can models automatically discover the hidden structures in the data and be trained directly from web data? Many people have pushed hard in this line of research. For example, earlier work focused on jointly modeling images and text and used text based search engines for gathering the data~\cite{berg2006animals,schroff2011harvesting,saenko2009unsupervised}. This tends to offer less biased training pairs, but unfortunately such an association is often too weak and hard to capture, since visual knowledge is usually regarded as common sense knowledge and too obvious to be mentioned in the text~\cite{chen2013neil}. As the image search engines became mature, recent work focused on using them to filter out the noise when learning visual models~\cite{fergus2004visual,wang2008annotating,vijayanarasimhan2008keywords,torresani2010efficient,li2013harvesting,divvala2014learning,golge2014conceptmap}. But using image search engines added more bias to the gathered data ~\cite{bergamo2010exploiting,mezuman2012learning,lin2014microsoft}. To combat both noise and data bias, recent approaches have taken a more semi-supervised approach. In particular, \cite{li2010optimol,chen2013neil} proposed iterative approaches to jointly learn models and find clean examples, hoping that simple examples learned first can help the model learn harder, more complex examples~\cite{bengio2009curriculum,kumar2010self}. However, to the best of our knowledge, human supervision is still a clear winner in performance, regardless of orders of magnitudes more data seen by many of these web learners.

Our work is also closely related to another trend in computer vision: learning and exploiting visual representation via CNNs~\cite{krizhevsky2012imagenet,girshick2014rich,taigman2014deepface,hariharan2014simultaneous}. However, learning these CNNs from noisy labeled data~\cite{sukhbaatar2014learning,reed2014training} is still an open challenge. Following the recent success of convolutional networks and curriculum learning~\cite{bengio2009curriculum,kumar2010self,lee2011learning}, we demonstrate that, while directly training CNNs with high-level or fine-grained queries (\eg random proper nouns, abstract concepts) and noisy labels (\eg Flickr tags) can still be challenging, a more learning approach might provide us the right solution. Specifically, one can bootstrap CNN training with easy examples first, followed by a more extensive and comprehensive learning procedure with similarity constraints to learn visual representations. We demonstrate that visual representations learned by our algorithm performs very competitively as compared to ImageNet trained CNNs.

Finally, our paper is also related to learning from weak or noisy labels~\cite{crandall2006weakly,pandey2011scene,deselaers2012weakly,song2014learning,wang2014weakly}. There are some recent works showcasing that CNNs trained in a weakly supervised setting can also develop detailed information about the object intrinsically~\cite{simonyan2013deep,Oquab13,pathak2014fully,bergamo2014self,papandreou2015weakly}. However, different from the assumptions in most weakly-supervised approaches, here our model is deprived of clean human supervision altogether (instead of only removing the location or segmentation). Most recently, novel loss layers have also been introduced in CNNs to deal with noisy labels~\cite{sukhbaatar2014learning,reed2014training}. On the other hand, we assume a vanilla CNN is robust to noise when trained with simple examples, from which a relationship graph can be learned, and this relationship graph provides powerful constraints when the network is faced with more challenging and noisier data. 

\vspace{-0.1in}
\section{Approach}
Our goal is to learn deep representations directly from the massive amount of data online. While it seems that CNNs are designed for big data - small datasets plus millions of parameters can easily lead to over-fitting, we found it is still hard to train a CNN naively with random image-text/tag pairs. For example, most Flickr tags correspond to meta information and specific locations, which usually results in extremely high intra-tag variation. One possibility is to use commercial text-based image search engine to increase diversity in the training data. But if thousands of query strings are used some of them might not correspond to a visualizable concept and some of the query strings might be too fine grained (\eg random names of a person or abstract concepts). These non-visualizable concepts and fine-grained categories incur unexpected noise during the training process\footnote{We tried to train a CNN with Google results of $\sim$7000 noun phrases randomly sampled from the web ($\sim$5M images), but it does not converge.}. One can use specifically designed techniques~\cite{chen2013neil,divvala2014learning} and loss layers~\cite{sukhbaatar2014learning,reed2014training} to alleviate some of these problems. But these approaches are based on estimating the empirical noise distribution which is non-trivial. Learning the noise distribution is non-trivial since it is heavily dependent on the representation, and weak features (\eg HOG or when the network is being trained from scratch) often lead to incorrect estimates. On the other hand, for many basic categories commonly used in the vision community, the top results returned by Google image search are pretty clean. In fact, they are so clean that they are biased towards iconic images where a single object is centered with a clean background in a canonical viewpoint~\cite{mezuman2012learning,raguram2008computing,berg2009finding,lin2014microsoft}. This is good news for learning algorithm to quickly grasp the appearance of a certain concept, but a representation learned from such data is likely biased and less generalizable. So, what we want is an approach that can learn visual representation from Flickr-like images.

Inspired by the philosophy of curriculum learning~\cite{bengio2009curriculum,kumar2010self,lee2011learning}, we take a two-step approach to train CNNs from the web. In curriculum learning, the model is designed to learn the easy examples first, and gradually adapt itself to harder examples. In a similar manner, we first train our CNN model from scratch using easy images downloaded from Google image search. Once we have this representation learned we try to feed harder Flickr images for training. Note that training with Flickr images is still difficult because of noise in the labels. Therefore, we apply constraints during fine-tuning with Flickr images. These constraints are based on similarity relationships across different categories. Specifically, we propose to learn a relationship graph and initial visual representation from the easy examples first, and later during fine-tuning, the error can back-propagate through the graph and get properly regularized. % To demonstrate the effectiveness of our representation, we do two experiments. 1) We use our trained CNNs for object detection on VOC 2007 and 2012, and scene classification on MIT Indoor-67. 2) We use the representation to clean up the noisy images and train R-CNN detectors directly from the web. These detectors are tested on standard VOC 2007 dataset. 
The outline of our approach is shown in Figure~\ref{fig:outline}.

\subsection{Initial Network}
As noted above, common categories used in vision nowadays are well-studied and search engines give relatively clean results. Therefore, instead of using random noun phrases, we obtained three lists of categories from ImageNet Challenge~\cite{russakovsky2014imagenet}, SUN database~\cite{xiao2010sun} and NEIL knowledge base~\cite{chen2013neil}. ImageNet syn-sets are transformed to its surface forms by just taking the first explanation, with most of them focusing on object categories. To better assist querying and reducing noise, we remove the suffix (usually correspond to attributes, \eg indoor/outdoor) of the SUN categories. Since NEIL is designed to query search engines, its list is comprehensive and favorable, we collected the list for objects and attributes and removed the duplicate queries with ImageNet. The category names are directly used to query Google for images. Apart from removing unreadable images, no pre-processing is performed. This leave us with $\sim$600 images for each query. All the images are then fed directly into the CNN as training data. 

For fair comparison, we use the same architecture (besides the output layer) as the BLVC reference network~\cite{jia2014caffe}, which is a slight variant of of the original network proposed by~\cite{krizhevsky2012imagenet}. The architecture has five convolutional layers followed by two fully connected layers. After seventh layer, another fully connected layer is used to predict class labels. 

\subsection{Representation Adaptation with Graph}
After converging, the initial network has already learned favorable low-level filters to represent the ``visual world'' outlined by Google image search. However, as mentioned before, this ``visual world'' is biased toward clean and simple images. For example, it was found that more than 40\% of the cars returned by Google are viewed from a 45 degree angle~\cite{mezuman2012learning}. Moreover, when a concept is a product, lots of the images are wallpapers and advertisements with artificial background, with the product centered and pictured from the best selling view. On the other hand, photo-sharing websites like Flickr have more realistic images since the users upload their own photos. Though photographic bias still exists, most of the images are closer-looking to the visual world humans experience everyday. Datasets constructed from them are shown to generalize better~\cite{torralba2011unbiased,lin2014microsoft}. Therefore, as a next step, we aim to narrow the gap by fine-tuning our representation on Flickr images~\footnote{Flickr images are downloaded using tag search. We use the same query strings as used in Google image search.}.

For fine-tuning the network with hard Flickr images, we again feed these images as-is for training, with the tags as class labels. While we are getting more realistic images, we did notice that the data becomes noisier. Powerful as CNNs, they are still likely to be diluted by the noisy examples over the fine-tuning process\footnote{In our experiments, we find with the same $\sim$1500 categories and close-to-uniform label distribution, a CNN converged on Google images yields an entropy $\sim$2.8, whereas Flickr gives $\sim$4.0. Note that complete random noise will give $\sim$$\log$(1500)=7.3 and perfectly separable signal close to 0.0.}. In an noisy open-domain environment, mistakes are unavoidable. But humans are more intelligent: we not just learn to recognize concepts independently, but also build up interconnections and develop theories to help better understand the world~\cite{carruthers1996theories}. Inspired by this, we want to train CNNs with such relationships - with their simplest form being pair-wise look-alike ones~\cite{chen2013neil,divvala2014learning}. Such a relationship graph can provide more information of the class and regularize/constrain the network training. A motivating example is ``iphone''. While Google mostly returns images of the product, on Flickr it is often used to specify the device a photo is taken with - as a result, virtually any image can be tagged as ``iphone''. Knowing similar-looking categories to ``iphone'' can intuitively help here.

One way to obtain relationships is through extra knowledge sources like WordNet~\cite{miller1995wordnet}. However, they are not necessarily developed for the visual domain. Instead, we take a data-driven approach to discover relationships in our data: we assume the network will intrinsically develop connections between different categories when clean examples are offered, and all we have to do is to distill the knowledge out.

We take a simple approach by just testing our network on the training set, and take the confusion matrix as the relationships. Mathematically, for any pair of concepts $i$ and $j$, the relationship $R_{ij}$ is defined as:
\begin{equation}
R_{ij} = P(i | j) = \frac{\sum_{k\in C_i}CNN(j|I_k)}{|C_i|},
\end{equation}
where $C_i$ is the set of indexes for images that belong to concept $i$, $|\cdot|$ is the cardinality function, and given pixel values $I_k$, $CNN(j|I_k)$ is the network's belief on how likely image $k$ belongs to concept $i$. We want our graph to be sparse, therefore we just used the top $K$ ($K=5$ in our experiments) and re-normalized the probability mass. 

After constructing the relationship graph, we put this graph (represented as a matrix) on top of the seventh layer of the network, so that now the soft-max loss function becomes:
\begin{equation}
L = \sum_{k}\sum_{i}R_{il_k}\log(CNN(i|I_k)),
\end{equation}
where $l_k$ is the class label.
In this way, the network is trained to predict the context of a category (in terms of relationships to other categories), and the error is back-propagated through the relationship graph to lower layers. Note that, this extra layer is similar to~\cite{sukhbaatar2014learning}, in which $R_{ij}$ is used to characterize the label-flip noise. Different from them, we do not assume all the categories are \emph{mutually exclusive}, but instead \emph{inter related}. For example, ``cat'' is a hyper-class of ``Siamese cat'', and it is reasonable if the model believes some examples of ``Siamese cat'' are more close to the average image of a ``cat''. Please see Section~\ref{sec:exp} for our empirical validation of this assumption. For fear of semantic drift, in this paper we keep the initially learned graph structure fixed, but it would be interesting to see how updating the relationship graph performs (like~\cite{chen2013neil}). 

\begin{figure}[t]
\centering
\includegraphics[width=0.97\linewidth]{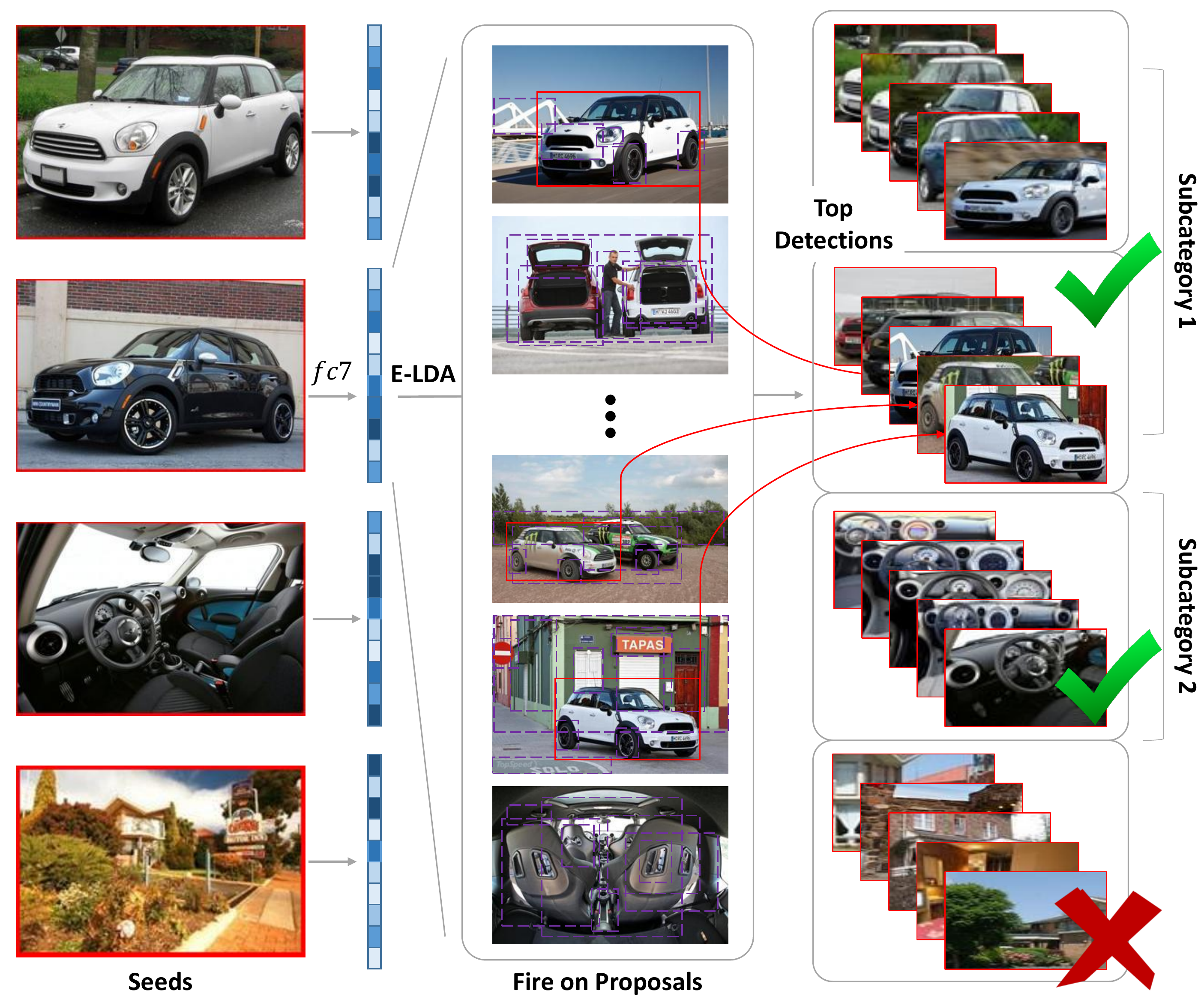}
\caption{{\small Our pipeline of object localization (for ``countryman''). E-LDA detectors~\cite{hariharan2012discriminative} trained on $fc7$ features of the seed images are fired on EdgeBox proposals (purple boxes) from other images for nearest neighbors (red boxes), which are then merged to form subcategories. Noisy subcategories are purged with density estimation~\cite{chen_cvpr14}. \label{fig:loc}}}
\vspace{-0.25in}
\end{figure}

\subsection{\label{sec:loc}Localizing Objects}
% To show the effectiveness of our representation, after fine-tuning we go back to the problem of organizing the data on the web: that is, clean up the data by removing noise and localizing objects in the images. But shouldn't the CNN have learned intrinsically the salient regions in an image for the concepts of interest~\cite{simonyan2013deep,bergamo2014self,papandreou2015weakly}? Isn't getting clean data as simple as ranking the initial set of images based on the soft-max output? We argue that, while the network has already learned to model the \emph{positive} examples when solving the multi-way classification problem, it has not yet learned the distribution of \emph{negative} data, \eg background clutter. While scenes and attributes are more ``stuff-like'' and thus finding clean full images might be enough, it is important for objects to be localized well, particularly when they are small in the original image. In fact, since the network is optimized for a classification loss, the representation is learned to be spatially invariant (\eg, the network should output ``orange'' regardless of where it exists in the image, and how many there are), precisely localizing the object is a very challenging task.
Until now, we have focused on learning a webly-supervised CNN representation based on classification loss. In order to train a webly-supervised object detector we still need to clean the web data and localize the objects in those images to train a detector like R-CNN~\cite{girshick2014rich}. Note that this is a non-trivial task, since: 1) the CNN is only trained to distinguish a closed set of classes, unnecessarily aware of all the negative visual world, \eg background clutter; 2) the classification loss encourages the representation to be spatially \emph{invariant} (\eg, the network should output ``orange'' regardless of where it exists in the image or how many there are), which can be a serious issue for localization.

We now describe our subcategory discovery based approach similar to~\cite{chen2013neil} to clean data and localize objects. The whole process is illustrated in Figure~\ref{fig:loc}.

\vspace{0.05in}
\noindent {\bf Seeds}: We use the full images returned by Google as seed bounding boxes. This is based on Google's bias toward images with a single centered object and a clean background.

\noindent {\bf Nearest Neighbor Propagation}: For each seed, we train an Exemplar-LDA~\cite{hariharan2012discriminative} detector using our trained $fc7$ features. Negative statistics for E-LDA are computed over all the downloaded images. This E-LDA detector is then fired on the remaining images to find its top $k$ nearest neighbors. For efficiency, instead of checking all possible windows on each image, we use EdgeBox~\cite{zitnick2014edge} to propose candidate ones, which also reduces background noise. We set $k$=10 in our experiments. 

\begin{figure*}[]
\centering
\includegraphics[width=0.97\linewidth]{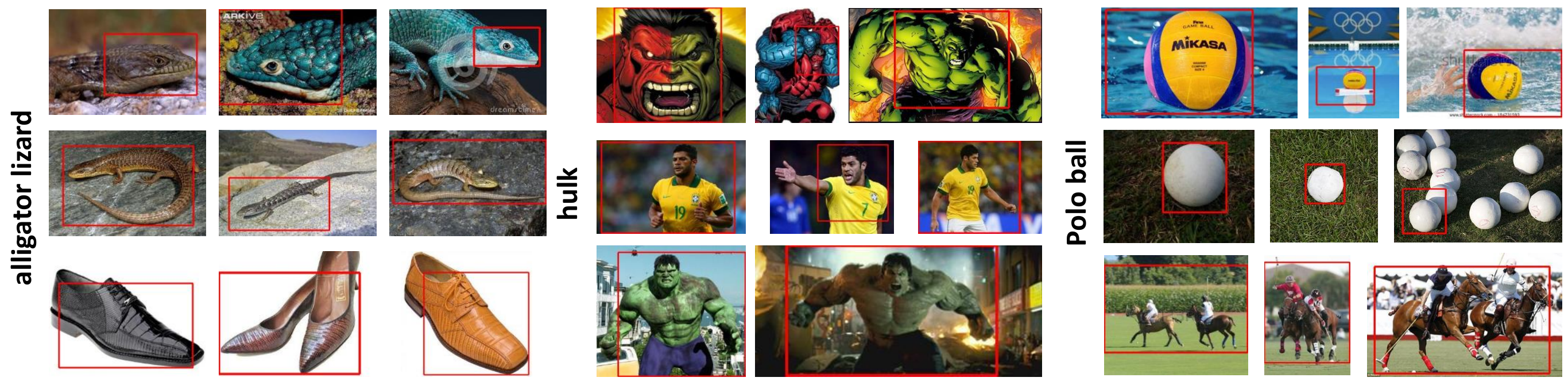}    
\caption{{\small We use the learned CNN representation to discover subcategories and localize positive instances for different categories~\cite{chen2013neil}.}\label{fig:subcat}}
\vspace{-0.2in}
\end{figure*}

\noindent {\bf Clustering into Subcategories}: We then use a publicly-available variant of agglomerative clustering~\cite{chen_cvpr14} where the nearest neighbor sets are merged iteratively from bottom up to form the final subcategories based on E-LDA similarity scores and density estimation. Note that this is different from~\cite{chen2013neil}, but gives similar results while being much more efficient. Some example subcategories are shown in Figure~\ref{fig:subcat}.

%\footnote{To further reduce the computation overhead, we only use windows that cover $\ge$ 1\% of the entire image. We find it only has negligible effect on the final clustering quality, but purges $\ge$ 90\% of the proposals.}

%\red{To show the effectiveness of our representation, after fine-tuning we go back to the problem of organizing the web data: that is, clean up the data by removing noise and localizing objects in the images. Note that this is a non-trivial task, since: 1) the CNN is only trained to distinguish a closed set of classes, unnecessarily aware of all the negative visual world, \eg background clutter; 2) the classification loss encourages the representation to be spatially \emph{invariant} (\eg, the network should output ``orange'' regardless of where it exists in the image or how many there are), which can be a serious issue for localization. }

\begin{figure*}[t]
\centering
\includegraphics[width=1.0\linewidth]{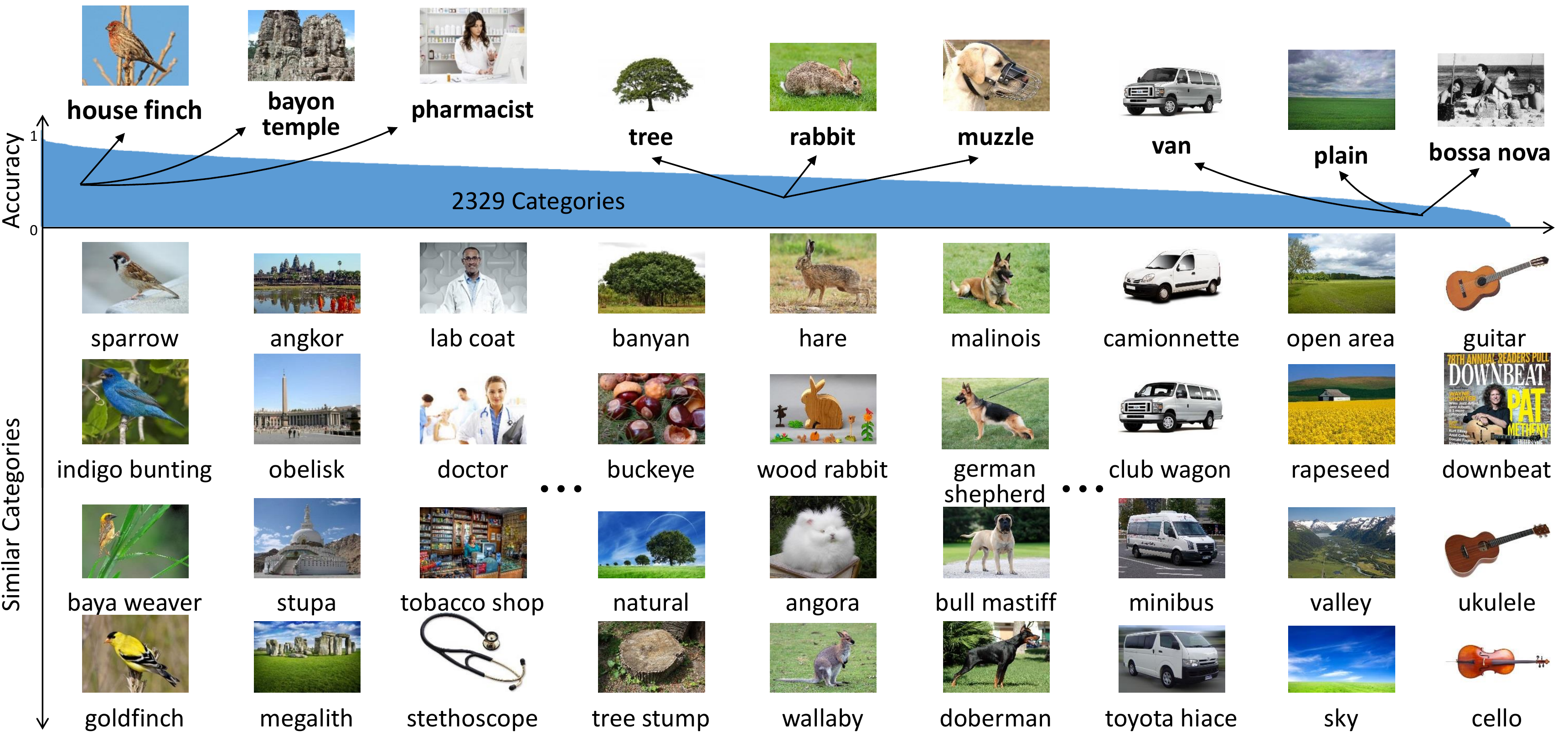}
\caption{{\small Visualization of the relationships learned from the confusion matrix. The horizontal axis is for categories, which are ranked based on CNN's accuracy. Here we show random examples from three parts of the distribution: top, middle, bottom. It can be seen that the relationships are reasonable: at the top of the distribution the network can recognize well, but when it gets confused, it gets confused to similar categories. Even for bottom ones where the network gets heavily confused, it is confusing between semantically related categories. Somewhat to our surprise, for noisy classes like ``bossa nova'', the network can figure out it is related to musical instruments.}\label{fig:rel}}
\vspace{-0.1in}
\end{figure*}

\vspace{0.05in}
Finally, we train a R-CNN~\cite{girshick2014rich} detector for each category based on all the clustered bounding boxes. Random patches from YFCC~\cite{yfcc} are used as negatives. The naive approach would be using the positive examples as-is. Typically, hundreds of instances per category are available for training. While this number is comparable to the VOC 2007 trainval set~\cite{everingham2010pascal}, we also tried to increase positive bounding boxes using two strategies:

%\begin{description}
%\setlength\itemsep{0em}
%\item[Edge-Box Augmentation (EA)] 
\vspace{0.05in}
\noindent {\bf EdgeBox Augmentation (EA)}: We follow~\cite{girshick2014rich} to augment the positive training examples. We again use EdgeBox~\cite{zitnick2014edge} to propose regions of interest on images. Whenever a proposal has a $\ge$0.5 overlapping (measured by intersection over union) with any of the positive bounding box, we add it for training.
%\item[Category Expansion (CE)] 

\noindent {\bf Category Expansion (CE)}: One big advantage of Internet is its nearly infinite data limit. Here we again use the relationship graph to look for similar categories for more training examples. After verification the semantic-relatedness with WordNet~\cite{miller1995wordnet}, we add the examples into training dataset. We believe the extra examples should allow better generalization.
%\end{description}

\vspace{0.05in}
Note both these strategies are only used to increase the amount of positive data for the final SVM to be trained in R-CNN. We do not re-train our CNN representations using these strategies.

\section{Experimental Results\label{sec:exp}}
We now describe our experimental results. Our goal is to demonstrate that the visual representation learned using two-step webly supervised learning is meaningful. For this, we will do four experiments: 1) First, we will show that our learned CNN can be used for object detection. Here, we use the approach similar to R-CNN~\cite{girshick2014rich} where we will fine-tune our learned CNN using VOC data. This is followed by learning SVM-detectors using CNN features. 2) We will also show that our CNN can be used to clean up the web data: that is, discover subcategories and localize the objects in web images. 3) We will train detectors using the cleaned up web data and evaluate them on VOC data. Note in this case, we will not use any VOC training images. We will only use web images to train both the CNN and the subsequent SVMs. 4) Finally, we will show scene classification results to further showcase the usefulness of the trained representation.

All the networks are trained with the Caffe Toolbox~\cite{jia2014caffe}. In total we have 2,240 objects, 89 attributes, and 874 scenes. Two networks are trained on Google: 1) The object-attribute network (GoogleO), where the output dimension is 2,329, and 2) All included network (GoogleA), where the output dimension is 3,203. For the first network, $\sim$1.5 million images are downloaded from Google image search. Combining scene images, $\sim$2.1 million images are used in the second network. We set the batch size as 256 and start with a learning rate of 0.01. With Xavier initialization~\cite{glorot2010understanding}, the learning rate is reduced by a factor of 10 after every 150K iterations, and we stop training at 450K iterations. For two-stage training, GoogleO is then fine-tuned with $\sim$1.2 million Flickr images. We tested both with (FlickrG) and without (FlickrF) the relationship graph as regularization. Fine-tuning is performed for a total of 100K iterations, with a step size of 30K. As baseline, we also report numbers for CNN learned using Flickr images alone (FlickrS) and combined Google+Flickr images (GFAll). Note in case of GFAll, neither two stage learning or relationship graph constraint is used.

\begin{table*}
\scriptsize{
\setlength{\tabcolsep}{3pt}
\def\arraystretch{1.2}
\center
\begin{tabular}{c c @{}l c c c c c c c c c c c c c c c c c c c c@{}p{0.3cm}@{}c@{}}
\toprule
& & \textbf{VOC 2007 test}  & aero  &   bike &  bird & boat &  bottle  &  bus  &  car  &  cat  &  chair & cow & table &  dog  & horse & mbike & pers  & plant & sheep & sofa & train & tv &  &\textbf{mAP}\\
\midrule
% \parbox[t]{2mm}{\multirow{6}{*}{\rotatebox[origin=c]{90}{w/o VOC FT}}} & & ImageNet~\cite{girshick2014rich} & 57.6 & 57.9 & 38.5 & 31.8 & 23.7 & 51.2 & 58.9 & 51.4 & 20.0 & 50.5 & 40.9 & 46.0 & 51.6 & 55.9 & 43.3 & 23.3 & 48.1 & 35.3 & 51.0 & 57.4 &  & 44.7 \\
% \cmidrule{3-25}
% & & GoogleO & 57.1 & 59.9 & 35.4 & 30.5 & 21.9 & 53.9 & 59.5 & 40.7 & 18.6 & 43.3 & 37.5 & 41.9 & 49.6 & 57.7 & 38.4 & 22.8 & 45.2 & 37.1 & 48.0 & 54.5 &  & 42.7 \\
% & & GoogleA  & 54.9 & 58.2 & 35.7 & 30.7 & 22.0 & 54.5 & 59.9 & 44.7 & 19.9 & 41.0 & 34.5 & 40.1 & 46.8 & 56.2 & 40.0 & 22.2 & 45.8 & 36.3 & 47.5 & 54.2 &  & 42.3 \\
% & & \red{FlickrS} & 50.0 & 55.9 & 29.6 & 26.8 & 18.7 & 47.6 & 56.3 & 34.4 & 14.5 & 35.9 & 33.3 & 34.2 & 43.2 & 52.2 & 36.7 & 21.5 & 43.3 & 31.6 & 48.5 & 48.4 &  & \red{38.1} \\
% & & \red{FlickrF} & 53.9 & 60.7 & 37.0 & 31.6 & 23.8 & 57.7 & 60.8 & 44.1 & 20.3 & 46.5 & 31.5 & 39.8 & 49.7 & 59.0 & 41.6 & 23.0 & 44.4 & 36.2 & 49.9 & 56.2 &  & \red{43.4} \\
% & & FlickrG  & 55.3 & 61.9 & 39.1 & 29.5 & 24.8 & 55.1 & 62.7 & 43.5 & 22.7 & 49.3 & 36.6 & 42.7 & 48.9 & 59.7 & 41.2 & 25.4 & 47.7 & 41.9 & 48.8 & 56.8 &  & 44.7 \\
% & & \red{GFALL} & 52.1 & 57.8 & 38.1 & 25.6 & 21.2 & 47.6 & 56.4 & 43.8 & 19.6 & 42.6 & 30.3 & 37.6 & 45.1 & 50.8 & 39.3 & 22.9 & 43.5 & 34.2 & 48.3 & 52.2 &  & \red{40.5} \\

\parbox[t]{1.5mm}{\multirow{7}{*}{\rotatebox[origin=c]{90}{w/o VOC FT}}} & & ImageNet~\cite{girshick2014rich} & \textbf{57.6} & 57.9 & 38.5 & \textbf{31.8} & 23.7 & 51.2 & 58.9 & \textbf{51.4} & 20.0 & \textbf{50.5} & \textbf{40.9} & \textbf{46.0} & \textbf{51.6} & 55.9 & \textbf{43.3} & 23.3 & \textbf{48.1} & 35.3 & \textbf{51.0} & \textbf{57.4} &  & \textbf{44.7} \\
\cmidrule{3-25}
& & GoogleO [Obj.] & 57.1 & 59.9 & 35.4 & 30.5 & 21.9 & 53.9 & 59.5 & 40.7 & 18.6 & 43.3 & 37.5 & 41.9 & 49.6 & 57.7 & 38.4 & 22.8 & 45.2 & 37.1 & 48.0 & 54.5 &  & 42.7 \\
& & GoogleA  [Obj. + Sce.] & 54.9 & 58.2 & 35.7 & 30.7 & 22.0 & 54.5 & 59.9 & 44.7 & 19.9 & 41.0 & 34.5 & 40.1 & 46.8 & 56.2 & 40.0 & 22.2 & 45.8 & 36.3 & 47.5 & 54.2 &  & 42.3 \\
& & FlickrS [Flickr Obj.] & 50.0 & 55.9 & 29.6 & 26.8 & 18.7 & 47.6 & 56.3 & 34.4 & 14.5 & 35.9 & 33.3 & 34.2 & 43.2 & 52.2 & 36.7 & 21.5 & 43.3 & 31.6 & 48.5 & 48.4 &  & 38.1 \\
& & GFAll [All Obj., 1-stage] & 52.1 & 57.8 & 38.1 & 25.6 & 21.2 & 47.6 & 56.4 & 43.8 & 19.6 & 42.6 & 30.3 & 37.6 & 45.1 & 50.8 & 39.3 & 22.9 & 43.5 & 34.2 & 48.3 & 52.2 &  & 40.5 \\
& & FlickrF [2-stage] & 53.9 & 60.7 & 37.0 & 31.6 & 23.8 & \textbf{57.7} & 60.8 & 44.1 & 20.3 & 46.5 & 31.5 & 39.8 & 49.7 & 59.0 & 41.6 & 23.0 & 44.4 & 36.2 & 49.9 & 56.2 &  & 43.4 \\%& & \red{FlickrF} & 56.4 & 59.4 & 38.6 & \textbf{33.9} & 23.9 & 53.7 & 61.3 & 47.3 & 21.3 & 48.8 & 38.6 & 42.6 & 50.0 & 60.8 & 41.0 & 23.8 & 46.7 & 37.8 & 48.6 & \textbf{57.8} &  & \red{44.6} \\
& & FlickrG [2-stage, Graph] & 55.3 & \textbf{61.9} & \textbf{39.1} & 29.5 & \textbf{24.8} & 55.1 & \textbf{62.7} & 43.5 & \textbf{22.7} & 49.3 & 36.6 & 42.7 & 48.9 & \textbf{59.7} & 41.2 & \textbf{25.4} & 47.7 & \textbf{41.9} & 48.8 & 56.8 &  & \textbf{44.7} \\

\toprule
\parbox[t]{1.5mm}{\multirow{5}{*}{\rotatebox[origin=c]{90}{w/ VOC FT}}} & & VOC-Scratch~\cite{agrawal2014analyzing} & 49.9 & 60.6 & 24.7 & 23.7 & 20.3 & 52.5 & 64.8 & 32.9 & 20.4 & 43.5 & 34.2 & 29.9 & 49.0 & 60.4 & 47.5 & 28.0 & 42.3 & 28.6 & 51.2 & 50.0 &  & 40.7 \\
& & ImageNet~\cite{girshick2014rich} & 64.2 & \textbf{69.7} & \textbf{50.0} & \textbf{41.9} & \textbf{32.0} & 62.6 & 71.0 & \textbf{60.7} & \textbf{32.7} & 58.5 & 46.5 & \textbf{56.1} & 60.6 & 66.8 & 54.2 & 31.5 & 52.8 & 48.9 & 57.9 & \textbf{64.7} &  & \textbf{54.2} \\
\cmidrule{3-25}
& & GoogleO  & \textbf{65.0} & 68.1 & 45.2 & 37.0 & 29.6 & 65.4 & 73.8 & 54.0 & 30.4 & 57.8 & 48.7 & 51.9 & \textbf{64.1} & 64.7 & 54.0 & \textbf{32.0} & \textbf{54.9} & 44.5 & 57.0 & 64.0 &  & 53.1 \\
& & GoogleA  & 64.2 & 68.3 & 42.7 & 38.7 & 26.5 & 65.1 & 72.4 & 50.7 & 28.5 & \textbf{60.9} & \textbf{48.8} & 51.2 & 60.2 & 65.5 & \textbf{54.5} & 31.1 & 50.5 & \textbf{48.5} & 56.3 & 60.3 &  & 52.3 \\
& & FlickrG  & 63.7 & 68.5 & 46.2 & 36.4 & 30.2 & \textbf{68.4} & \textbf{73.9} & 56.9 & 31.4 & 59.1 & 46.7 & 52.4 & 61.5 & \textbf{69.2} & 53.6 & 31.6 & 53.8 & 44.5 & \textbf{58.1} & 59.6 &  & 53.3 \\
\bottomrule
\end{tabular}
\vspace{3pt}
\caption{Results on VOC 2007 (PASCAL data used). Please see Section~\ref{sec:det} for more details. \label{tab:voc_2007}}
}
\vspace{-0.1cm}
\end{table*}

\begin{table*}
\scriptsize{
\setlength{\tabcolsep}{3pt}
\def\arraystretch{1.2}
\center
\begin{tabular}{c c @{}l c c c c c c c c c c c c c c c c c c c c@{}p{0.3cm}@{}c@{}}
\toprule
& & \textbf{VOC 2012 test}  & aero  &   bike &  bird & boat &  bottle  &  bus  &  car  &  cat  &  chair & cow & table &  dog  & horse & mbike & person  & plant & sheep & sofa & train & tv &  &\textbf{mAP}\\
\midrule
\parbox[t]{2mm}{\multirow{4}{*}{\rotatebox[origin=c]{90}{w/ VOC FT}}} & & ImageNet~\cite{girshick2014rich} & 68.1 & 63.8 & 46.1 & 29.4 & 27.9 & 56.6 & 57.0 & 65.9 & 26.5 & 48.7 & 39.5   & \textbf{66.2} & 57.3 & 65.4 & 53.2 & 26.2 & 54.5 & 38.1 & 50.6 & 51.6 &  & 49.6 \\
& & ImageNet-TV & \textbf{73.3} & 67.1 & 46.3 & 31.7 & 30.6 & 59.4 & 61.0 & \textbf{67.9} & 27.3 & \textbf{53.1} & 39.1 & 64.1 & \textbf{60.5} & 70.9 & 57.2 & 26.1 & \textbf{59.0} & 40.1 & 56.2 & \textbf{54.9} &  & 52.3 \\
\cmidrule{3-25}
& & GoogleO  & 72.2 & 67.3 & 46.0 & \textbf{32.3} & \textbf{31.6} & \textbf{62.6} & 62.5 & 66.5 & 27.3 & 52.1 & 38.9 & 64.0 & 59.1 & 71.6 & 58.0 & 27.2 & 57.6 & 41.3 & 56.3 & 53.7 &  & 52.4 \\
& & FlickrG  & 72.7 & \textbf{68.2} & \textbf{47.3} & 32.2 & 30.6 & 62.3 & \textbf{62.6} & 65.9 & \textbf{28.1} & 52.2 & \textbf{39.5} & 65.1 & 60.0 & \textbf{71.7} & \textbf{58.2} & \textbf{27.3} & 58.0 & \textbf{41.5} & \textbf{57.2} & 53.8 &  & \textbf{52.7} \\
\bottomrule
\end{tabular}
\vspace{3pt}
\caption{Results on VOC 2012. Since~\cite{girshick2014rich} only fine-tuned on the train set, we also report results on trainval (ImageNet-TV) for fairness. \label{tab:voc_2012}}
}
\vspace{-0.2in}
\end{table*}

\vspace{0.05in}
\noindent \textbf{Is Confusion Matrix Informative for Relationships?}
We first want to show if  the network has learned to discover the look-alike relationships between concepts in the confusion matrix. To verify the quality of the network, we take the GoogleO net and visualize the top-5 most confusing concepts (including self) to some of the categories. To ensure our selection has a good coverage, we first rank the diagonal of the confusing matrix (accuracy) in the descending order. Then we randomly sample 3 categories from the top-100, bottom-100, and middle-100 from the list. The visualization and explanations can be found in Figure~\ref{fig:rel}. We can see that the top relationships learned are indeed reasonable.

\begin{table*}
\scriptsize{
\setlength{\tabcolsep}{3pt}
\def\arraystretch{1.2}
\center
\begin{tabular}{@{}l c c c c c c c c c c c c c c c c c c c c@{}p{0.3cm}@{}c@{}}
\toprule
\textbf{VOC 2007 test}  & aero  &   bike &  bird & boat &  bottle  &  bus  &  car  &  cat  &  chair & cow & table &  dog  & horse & mbike & person  & plant & sheep & sofa & train & tv &  &\textbf{mAP}\\
\midrule
LEVAN~\cite{divvala2014learning} & 14.0 & 36.2 & 12.5 & 10.3 & 9.2 & 35.0 & \textbf{35.9} & 8.4 & \textbf{10.0} & 17.5 & 6.5 & 12.9 & \textbf{30.6} & 27.5 & 6.0 & 1.5 & 18.8 & 10.3 & 23.5 & 16.4 &  & 17.1 \\
\midrule
GoogleO & 30.2 & 34.3 & 16.7 & 13.3 & 6.1 & 43.6 & 27.4 & 22.6 & 6.9 & 16.4 & 10.0 & 21.3 & 25.0 & 35.9 & 7.6 & 9.3 & 21.8 & 17.3 & 31.0 & 18.1 &  & 20.7 \\
GoogleA & 29.5 & 38.3 & 15.1 & 14.0 & 9.1 & 44.3 & 29.3 & 24.9 & 6.9 & 15.8 & 9.7 & 22.6 & 23.5 & 34.3 & 9.7 & 12.7 & 21.4 & 15.8 & 33.4 & 19.4 &  & 21.5 \\
\midrule
FlickrG & 32.6 & 42.8 & 19.3 & 13.9 & 9.2 & 46.6 & 29.6 & 20.6 & 6.8 & 17.8 & 10.2 & 22.4 & 26.7 & 40.8 & 11.7 & \textbf{14.0} & 19.0 & 19.0 & 34.0 & 21.9 &  & 22.9 \\
FlickrG-EA & \textbf{32.7} & \textbf{44.3} & 17.9 & 14.0 & \textbf{9.3} & \textbf{47.1} & 26.6 & 19.2 & 8.2 & 18.3 & 10.0 & 22.7 & 25.0 & 42.5 & 12.0 & 12.7 & \textbf{22.2} & 20.9 & 35.6 & 18.2 &  & 23.0 \\
FlickrG-CE & 30.2 & 41.3 & \textbf{21.7} & \textbf{18.3} & 9.2 & 44.3 & 32.2 & \textbf{25.5} & 9.8 & \textbf{21.5} & \textbf{10.4} & \textbf{26.7} & 27.3 & \textbf{42.8} & \textbf{12.6} & 13.3 & 20.4 & \textbf{20.9} & \textbf{36.2} & \textbf{22.8} &  & \textbf{24.4} \\
\bottomrule
\end{tabular}
\vspace{3pt}
\caption{Webly supervised VOC 2007 detection results (No PASCAL data used). Please see Section~\ref{sec:subcat} for more details.\label{tab:uns}}
}
\vspace{-0.1in}
\end{table*}\textbf{}

\vspace{-0.2in}
\subsection{PASCAL VOC Object Detection\label{sec:det}}
Next, we test our webly trained CNN model for object detection on the PASCAL VOC. Following the R-CNN pipeline, two sets of experiments are performed on VOC 2007. First, we directly test the generalizability of CNN-representations learned without fine-tuning on VOC data. Second, we fine tune the CNN by back-propagating the error end-to-end using PASCAL trainval set. The fine-tuning procedure is performed 100K iteration, with a step size of 20K. In both cases, $fc7$ features are extracted to represent patches, and a SVM is learned to produce the final score.
%For fair comparison, no hyper parameter (\eg $C$) is tuned for R-CNN, keeping the settings identical to those for ImageNet.}

%Since we have trained various networks as potential candidates for PASCAL fine-tuning, we report numbers on all the networks trained. Note that GoogleO and GoogleA are trained with Google data, FlickrS is trained with Flickr data, but FlickrF, FlickrG and GFALL are trained with all the downloaded web data. The results are indicated in Table~\ref{tab:voc_2007}. 

%\footnote{\red{In fact, we also experimented training a separate CNN using Google images downloaded before March 2013. Despite having less data ($\sim$450 per category), the performance gap is only $\sim$1\% in the same settings (\eg categories used, learning strategy, \etc).} }

\begin{figure*}
\centering
\includegraphics[width=0.24\linewidth]{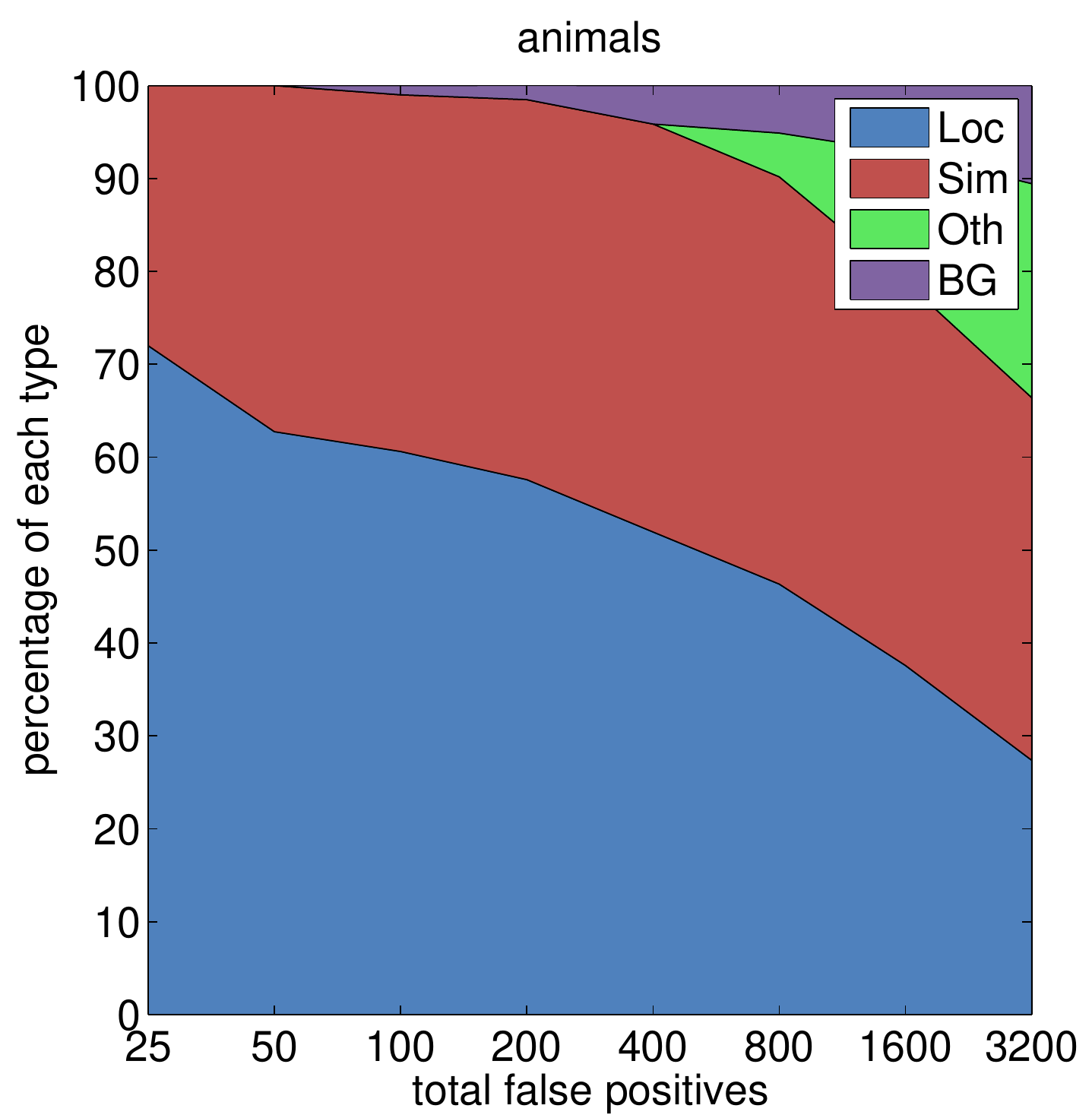}
\includegraphics[width=0.24\linewidth]{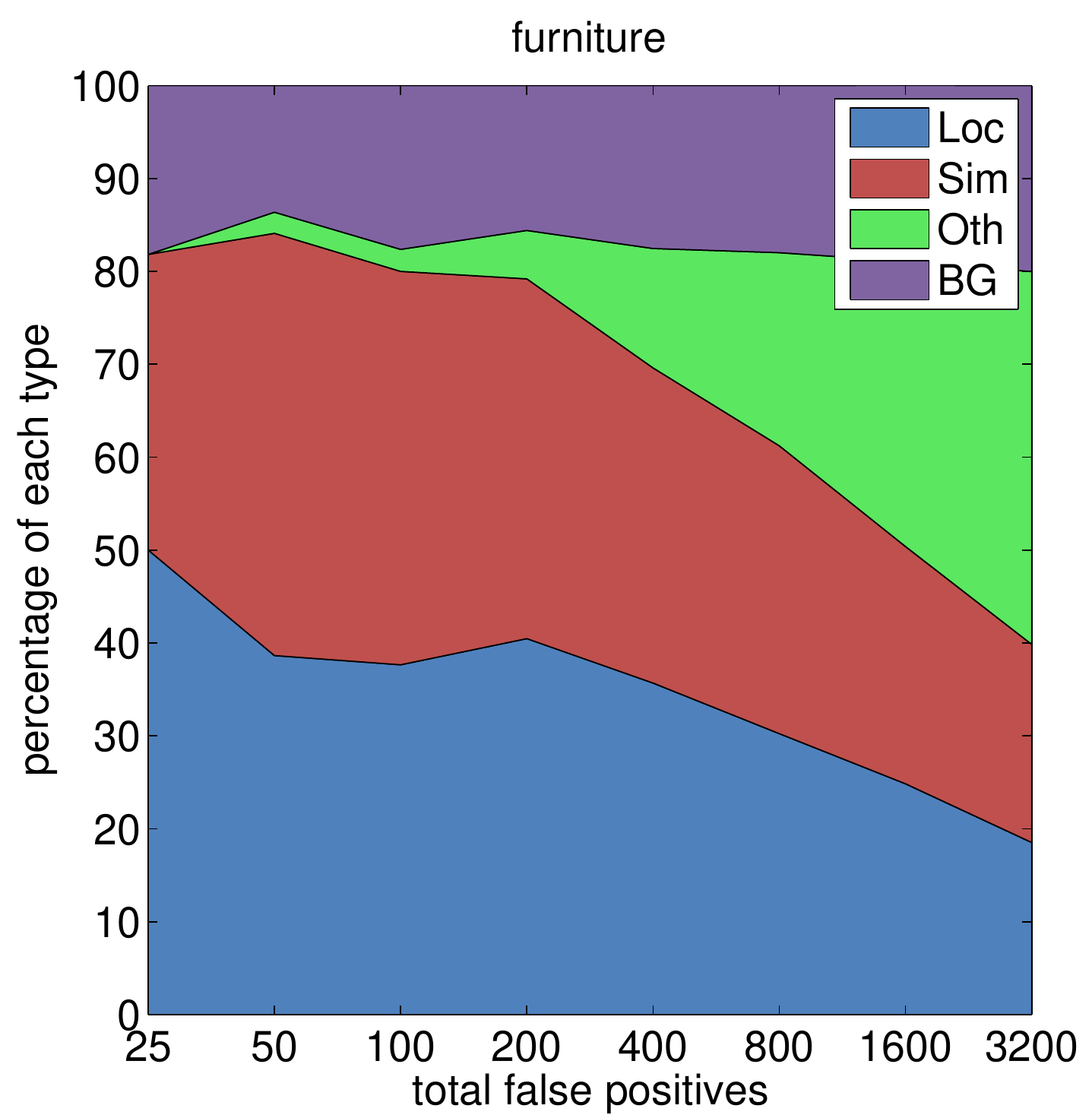}
\includegraphics[width=0.24\linewidth]{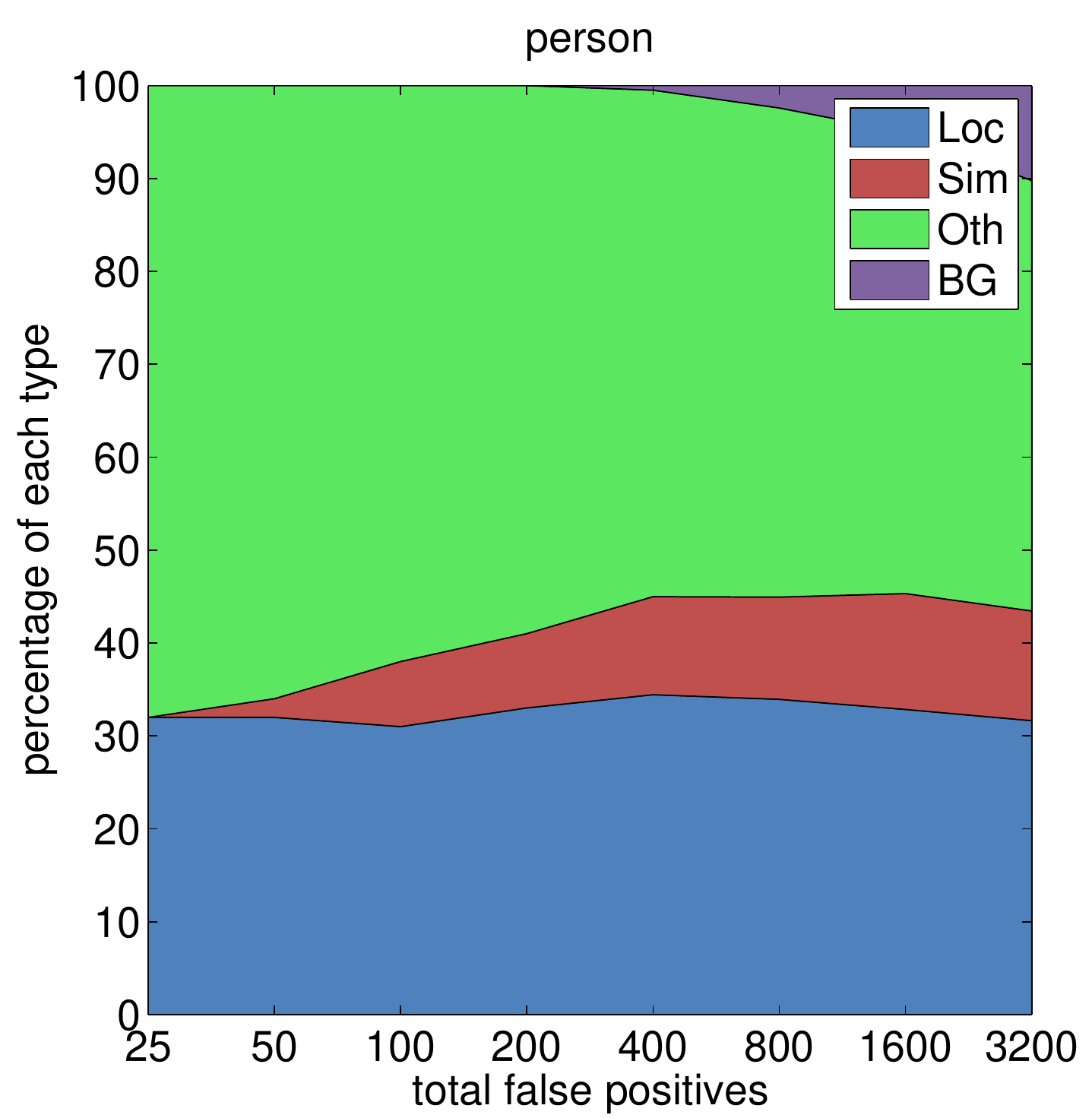}
\includegraphics[width=0.24\linewidth]{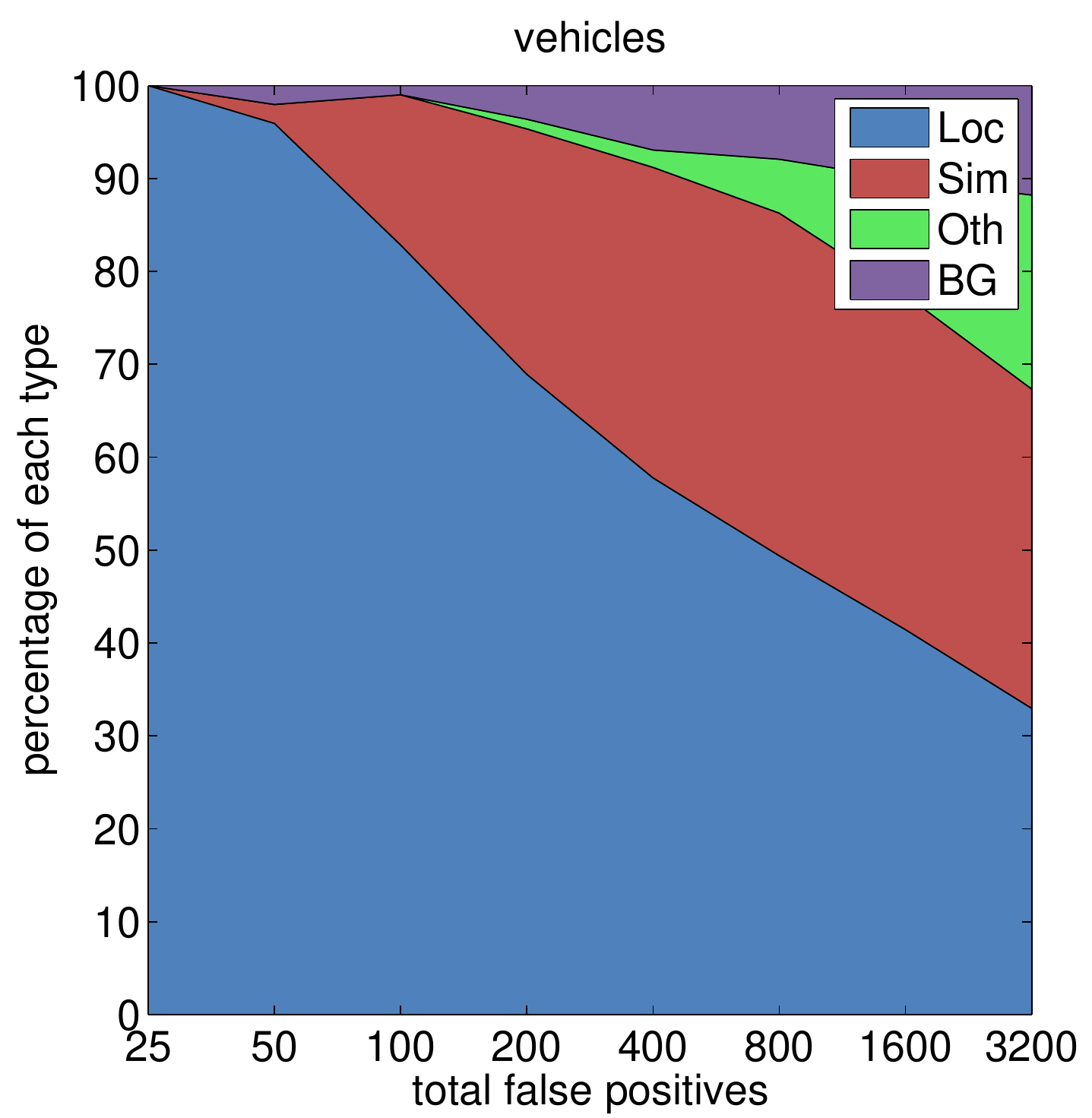}
\caption{{\small Diagnosis analysis using~\cite{hoiem2012diagnosing} for better understanding of the failure modes of our webly supervised pipeline. Please see top false positives in Figure~\ref{fig:fp}.}\label{fig:diag}}
\vspace{-0.1in}
\end{figure*}

We report numbers for all the CNNs on VOC 2007 data in Table~\ref{tab:voc_2007}. Several interesting notes:

\vspace{-0.1in}
\begin{itemize}
	\setlength\itemsep{0em}
	\item Despite the search engine bias and the noise in the data, our two-stage CNN with graph regularization is on par with ImageNet-trained CNN.
	\item Training a network directly on noisy and hard Flickr images hurt the learning process. For example, FlickrS gives the worst performance and in fact when a CNN is trained using all the images from Google and Flickr it gives a mAP of 40.5, which is substantially lower than our mAP.
	\item The proposed two-stage training strategy effectively takes advantage of the more realistic data Flickr provides. Without graph regularization we achieve a mAP of 43.4 (FlickrF). However, adding the graph regularization brings our final FlickrG network on par with ImageNet (mAP = 44.7).
%Despite the search engine bias toward simple images, Google images can be used to directly train CNNs}. 
%\item \red{Noisy Flickr images hurt representation. FlickrS not only gives the worst mAP, but also drags GoogleO's down when mixed with Google images (GFALL).}
\end{itemize}

We use the same CNNs for VOC 2012 and report results in Table~\ref{tab:voc_2012}. In this case, our networks outperform the ImageNet pretrained network even after fine-tuning (200K iterations, 40K step size). Note that the original R-CNN paper fine-tuned the ImageNet CNN using train data alone and therefore reports lower performance~\cite{girshick2014rich}. For fairness, we fine-tuned both ImageNet network and our networks on combined trainval images (ImageNet-TV). In both VOC 2007 and 2012, our webly supervised CNNs tend to work better for vehicles, probably because we have lots of data for cars and other vehicles ($\sim$500 classes). On the other hand, ImageNet CNN seems to outperform our network on animals~\cite{russakovsky2014imagenet} (\eg cat). This is probably because ImageNet has a lot more data for animals. It also suggests our CNNs can potentially benefit from more animal categories. 

\vspace{0.05in}
\noindent {\bf Does web supervision work because the image search engine is CNN-based?}
One possible hypothesis can be that our approach performs comparably to ImageNet-CNN because Google image search itself uses a trained CNN. To test if this hypothesis is true, we trained a separate CNN using NEIL images downloaded from Google before March 2013 (pre-CNN based image search era). Despite the data being noisier and less ($\sim$450 per category), we observe $\sim$1\% performance fall compared to a CNN trained with November 2014 data on the same categories. This indicates that the underlying CNN in Google image search has minimal effect on the training procedure and our network is quite robust to noise.

\subsection{Object Localization\label{sec:subcat}}
In this subsection, we are interested to see if we can detect objects without using a single PASCAL training image. We believe this is possible since we can localize objects automatically in web images with our proposed approach (see Section~\ref{sec:loc}). Please refer to Figure~\ref{fig:subcat} for the qualitative results on the training localization we can get with $fc7$ features. Compared to~\cite{chen2013neil}, the subcategories we get are less homogeneous (\eg people are not well-aligned, objects in different view points are clustered together). But just because of this more powerful representation (and thus better distance metric), we are able to dig out more signal from the training set - since semantically related images can form clusters and won't be purged as noise when an image is evaluated by its nearest neighbors. 

% \begin{figure*}[t]
% \centering
% \includegraphics[width=0.95\linewidth]{fp1}
% \caption{{\small Top false positives for selected categories on PASCAL VOC 2007 detection with Flickr-C. From top down: aeroplane, bicycle, bottle, cat.}\label{fig:fp}}
% \vspace{-0.2in}
% \end{figure*}

% \begin{figure*}[t]
% \centering
% \includegraphics[width=0.95\linewidth]{fp2}
% \caption{{\small Top false positives for selected categories on PASCAL VOC 2007 detection with Flickr-C. From top down: dining table, horse, person, tv monitor.}\label{fig:fp2}}
% \vspace{-0.2in}
% \end{figure*}

Using localized objects, we train R-CNN based detectors to detect objects on the VOC 2007 test set. We compare our results against~\cite{divvala2014learning}, who used Google $n$-grams to expand the categories (\eg ``horse'' is expanded to ``jumping horse'', ``racing horse'' \etc) and the models were also directly trained from the web. The results are shown in Table~\ref{tab:uns}. For our approach, we try five different settings: 1) GoogleO: Features are based on GoogleO CNN and the bounding boxes are also extracted only on easy Google images; 2) GoogleA: Using GoogleO to extract features instead; 3) FlickrG: Features are based on FlickrG instead; 4) FlickrG-EA: The same Flickr features are used but with EdgeBox augmentation; 5) FlickrG-CE: The Flickr features are used but the positive data includes examples from both original and expanded categories. From the results, we can see that in all cases the CNN based detector boosts the performance a lot. 

This demonstrates that our framework could be a powerful way to learn detectors for arbitrary object categories without labeling any training images. We plan to release a service for everyone to train R-CNN detectors on the fly. The code will also be released.

\begin{figure*}[t]
\centering
\includegraphics[width=0.95\linewidth]{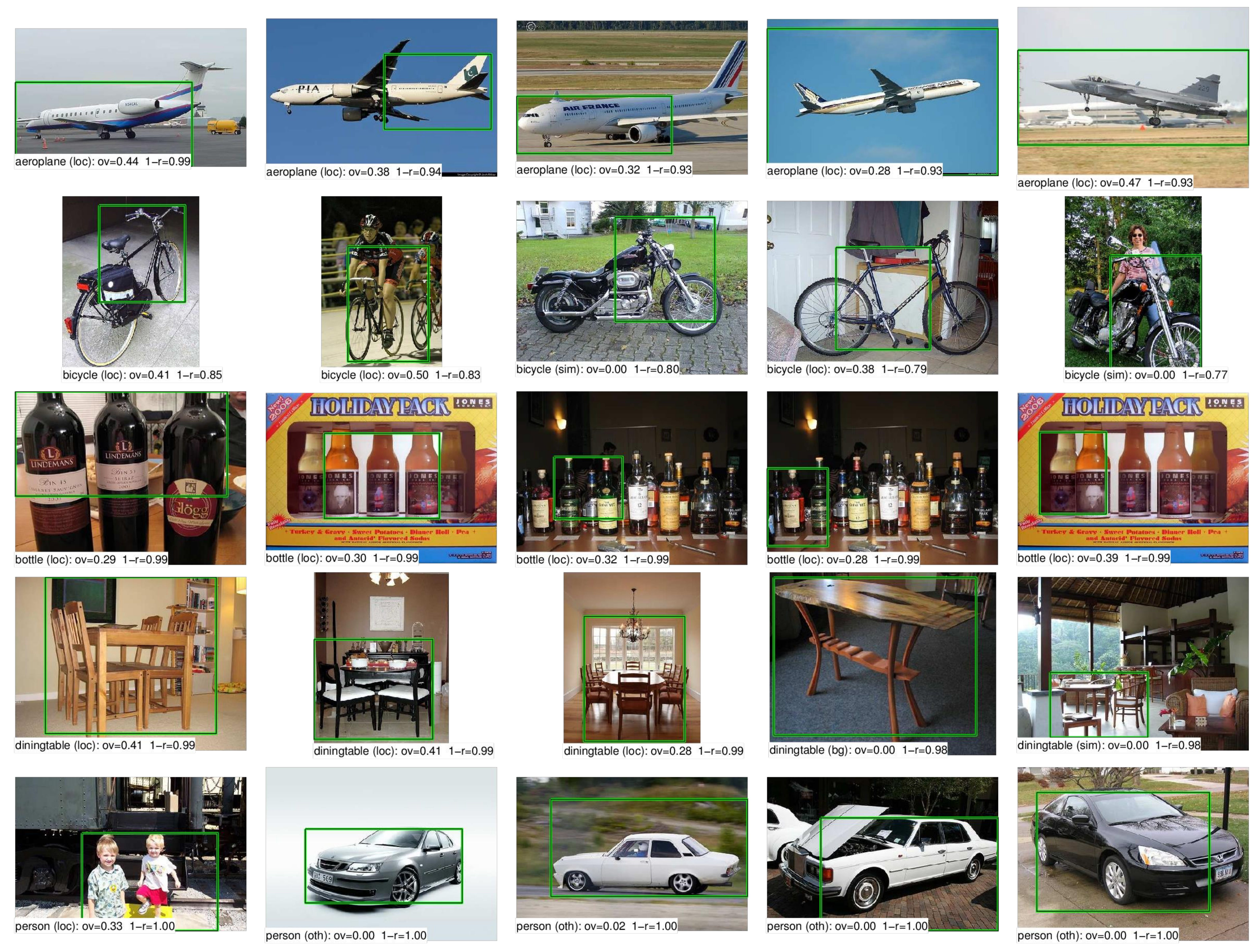}
\caption{{\small Top false positives for selected categories on PASCAL VOC 2007 detection with Flickr-C. From top down: aeroplane, bicycle, bottle, dinning table, and person.}\label{fig:fp}}
\vspace{-0.2in}
\end{figure*}

\subsection{Failure Modes for Webly Trained Detectors\label{sec:diag}}
In this section, we would like to gain more insights about the potential issues of our webly supervised object detection pipeline. We took the results from our best model (Flickr-C) and fed them to the publicly available diagnosis tool~\cite{hoiem2012diagnosing}. Figure~\ref{fig:diag} and \ref{fig:fp} highlight some of the interesting observations we found. 

Firstly, localization error accounts for a majority of the false positives. Since Google Image Search do not provide precise location information, the background is inevitably included when the detector is trained (\eg aeroplane, dining table). Multiple instances of an object can also occur in the image, but the algorithm has no clue that they should be treated as separate pieces (\eg bottle). Moreover, since our CNN is directly trained on full images, the objective function also biases the representation to be invariant (to spatial locations, \etc). All these factors caused localization issues.

Second, we did observe some interesting semantic drift between PASCAL categories and Google categories. For example, bicycle can also mean motorcycle on Google. Sense disambiguation for this polysemous word~\cite{saenko2009unsupervised,chen2015} is needed here. Also note that our person detector is confused with cars, we suspect it is because ``caprice'' was added as a related category but it can also mean a car (``chevy caprice''). How to handle such issues is a future research topic. 

%% TEMPORARY REMOVED
% \subsection{Failure Modes for Webly Trained Detectors\label{sec:diag}}
% In this subsection, we would like to gain more insights about the potential issues of our webly supervised object detection pipeline. We took the results from our best model (Flickr-C) and fed them to the publicly available diagnosis tool~\cite{hoiem2012diagnosing}. Figure~\ref{fig:diag} and \ref{fig:fp} highlight some of the interesting observations we found. 

% Firstly, localization error accounts for a majority of the false positives. Since Google Image Search do not provide precise location information, the background is inevitably included when the detector is trained (\eg aeroplane, dining table). Multiple instances of an object can also occur in the image, but the algorithm has no clue that they should be treated as separate pieces (\eg bottle). Moreover, since our CNN is directly trained on full images, the objective function also biases the representation to be invariant (to spatial locations, \etc). All these factors caused localization issues.

% Second, we did observe some interesting semantic drift between PASCAL categories and Google categories. For example, bicycle can also mean motorcycle on Google. Sense disambiguation for this polysemous word~\cite{saenko2009unsupervised,chen2015} is needed here. Also note that our person detector is confused with cars, we suspect it is because ``caprice'' was added as a related category but it can also mean a car (``chevy caprice''). How to handle such issues is a future research topic. 

\vspace{-0.1in}
\subsection{Scene Classification}
\vspace{-0.05in}

To further demonstrate the usage of CNN features directly learned from the web, we also conducted scene classification experiments on the MIT Indoor-67 dataset~\cite{quattoni2009recognizing}. For each image, we simply computed the $fc7$ feature vector, which has 4096 dimensions. We did not use any data augmentation or spatial pooling technique, with the only pre-processing step being normalizing the feature vector to unit $\ell_2$ length~\cite{razavian2014cnn}. The default SVM parameters ($C$=1) were fixed throughout the experiments. 

Table~\ref{tab:mit67} summarizes the results on the default train/test split. We can see our web based CNNs achieved very competitive performances: all the three networks achieved an accuracy at least on par with ImageNet pretrained models. Fine-tuning on hard images enhanced the features, but adding scene-related categories gave a huge boost to 66.5 (comparable to the CNN trained on Places database~\cite{zhou2014learning}, 68.2). This indicates CNN features learned directly from the web are generic and quite powerful.

Moreover, since we can easily get images for semantic labels (\eg actions, $n$-grams, \etc) other than objects or scenes from the web, webly supervised CNN bears a great potential to perform well on many relevant tasks - with the cost as low as providing a query list for that domain.

\begin{table}
\setlength{\tabcolsep}{30pt}
\def\arraystretch{1.2}
\center
\scriptsize{
\begin{tabular}{l c}
\toprule
\textbf{Indoor-67} & \textbf{Accuracy} \\
\midrule
ImageNet~\cite{zhou2014learning} & 56.8 \\
OverFeat~\cite{razavian2014cnn} & 58.4 \\
\midrule
GoogleO [Obj.] & 58.1 \\
FlickrG [Obj.] & 59.2 \\
GoogleA [Obj. + Sce.] & \textbf{66.5} \\
\bottomrule
\end{tabular}
\vspace{3pt}
\caption{Scene classification results on MIT Indoor-67. Note that GoogleA has scene categories for training but others do not.\label{tab:mit67}}
}
\vspace{-0.2in}
\end{table}

% \begin{figure*}[t]
% \centering
% \includegraphics[width=0.95\linewidth]{fp}
% \caption{{\small Top false positives for selected categories on PASCAL VOC 2007 detection with Flickr-C. From top down: aeroplane, bicycle, bottle, dinning table, and person.}\label{fig:fp}}
% \vspace{-0.2in}
% \end{figure*}

\vspace{-0.1in}
\section{Conclusion}
\vspace{-0.05in}
We have presented a two-stage approach to train CNNs using noisy web data. First, we train CNN with easy images downloaded from Google image search. This network is then used to discover structure in the data in terms of similarity relationships. Then we fine-tune the original network on more realistic Flickr images with the relationship graph. We show that our two-stage CNN comes close to the ImageNet pretrained-CNN on VOC 2007, and outperforms on VOC 2012. We report state-of-the-art performance on VOC 2007 without using any VOC training image. Finally, we will like to differentiate webly supervised and unsupervised learning. Webly supervised learning is suited for semantic tasks such as detection, classification (since supervision comes from text). On the other hand, unsupervised learning is useful for generic tasks which might not require semantic invariance (\eg, 3D understanding, grasping).

\vspace{-0.1in}
\section*{Acknowledgments}
\vspace{-0.05in}
This research is supported by ONR MURI N000141010934, Yahoo-CMU InMind program and a gift from Google. AG and XC were partially supported by Bosch Young Faculty Fellowship and Yahoo Fellowship respectively. The authors would also like to thank Yahoo! for a computing cluster and Nvidia for Tesla GPUs.

%We would like to emphasize that our CNN was trained with minimum explicit human supervision. We even show that our representation is so powerful that we can use it to organize the web data and learn arbitrary category detectors directly from the web. To the best of our knowledge, we show the best performance on VOC 2007 where no VOC data is used. \red{Additional results on scene understanding further demonstrate the effectiveness of our webly learned representation.}

% \vspace{-0.1in}
% \section*{Acknowledgements}
% This research is supported by ONR MURI N000141010934, Yahoo-CMU InMind program and a gift from Google. AG and XC were partially supported by Bosch Young Faculty Fellowship and Yahoo Fellowship respectively. The authors would also like to thank Yahoo! for the donation of a computing cluster and NVIDIA for Tesla K40 GPUs.

% \setstretch{0.91}

\vspace{-0.1in}
% {\footnotesize
% \bibliographystyle{ieee}
% \bibliography{egbib}

\begin{thebibliography}{10}\itemsep=-1pt

\bibitem{yfcc}
{YFCC} dataset.
\newblock \url{labs.yahoo.com/news/yfcc100m/}.

\bibitem{agrawal2014analyzing}
P.~Agrawal, R.~Girshick, and J.~Malik.
\newblock Analyzing the performance of multilayer neural networks for object
  recognition.
\newblock In {\em ECCV}. 2014.

\bibitem{bengio2009curriculum}
Y.~Bengio, J.~Louradour, R.~Collobert, and J.~Weston.
\newblock Curriculum learning.
\newblock In {\em ICML}, 2009.

\bibitem{berg2009finding}
T.~L. Berg and A.~C. Berg.
\newblock Finding iconic images.
\newblock In {\em CVPRW}, 2009.

\bibitem{berg2006animals}
T.~L. Berg and D.~A. Forsyth.
\newblock Animals on the web.
\newblock In {\em CVPR}, 2006.

\bibitem{bergamo2014self}
A.~Bergamo, L.~Bazzani, D.~Anguelov, and L.~Torresani.
\newblock Self-taught object localization with deep networks.
\newblock {\em arXiv:1409.3964}, 2014.

\bibitem{bergamo2010exploiting}
A.~Bergamo and L.~Torresani.
\newblock Exploiting weakly-labeled web images to improve object
  classification: a domain adaptation approach.
\newblock In {\em NIPS}, 2010.

\bibitem{carruthers1996theories}
P.~Carruthers and P.~K. Smith.
\newblock {\em Theories of theories of mind}.
\newblock Cambridge Univ Press, 1996.

\bibitem{chen2015}
X.~Chen, A.~Ritter, A.~Gupta, and T.~Mitchell.
\newblock Sense discovery via co-clustering on images and text.
\newblock In {\em CVPR}, 2015.

\bibitem{chen2013neil}
X.~Chen, A.~Shrivastava, and A.~Gupta.
\newblock {NEIL}: Extracting visual knowledge from web data.
\newblock In {\em ICCV}, 2013.

\bibitem{chen_cvpr14}
X.~Chen, A.~Shrivastava, and A.~Gupta.
\newblock Enriching visual knowledge bases via object discovery and
  segmentation.
\newblock In {\em CVPR}, 2014.

\bibitem{crandall2006weakly}
D.~J. Crandall and D.~P. Huttenlocher.
\newblock Weakly supervised learning of part-based spatial models for visual
  object recognition.
\newblock In {\em ECCV}. 2006.

\bibitem{deselaers2012weakly}
T.~Deselaers, B.~Alexe, and V.~Ferrari.
\newblock Weakly supervised localization and learning with generic knowledge.
\newblock {\em IJCV}, 2012.

\bibitem{divvala2014learning}
S.~K. Divvala, A.~Farhadi, and C.~Guestrin.
\newblock Learning everything about anything: Webly-supervised visual concept
  learning.
\newblock In {\em CVPR}, 2014.

\bibitem{everingham2010pascal}
M.~Everingham, L.~VanGool, C.~Williams, J.~Winn, and A.~Zisserman.
\newblock The pascal visual object classes (voc) challenge.
\newblock {\em IJCV 10}.

\bibitem{fan2010harvesting}
J.~Fan, Y.~Shen, N.~Zhou, and Y.~Gao.
\newblock Harvesting large-scale weakly-tagged image databases from the web.
\newblock In {\em CVPR}, 2010.

\bibitem{lsvm-pami}
P.~F. Felzenszwalb, R.~B. Girshick, D.~McAllester, and D.~Ramanan.
\newblock Object detection with discriminatively trained part based models.
\newblock {\em TPAMI}, 2010.

\bibitem{fergus2010learning}
R.~Fergus, L.~Fei-Fei, P.~Perona, and A.~Zisserman.
\newblock Learning object categories from internet image searches.
\newblock {\em Proceedings of the IEEE}, 2010.

\bibitem{fergus2004visual}
R.~Fergus, P.~Perona, and A.~Zisserman.
\newblock A visual category filter for google images.
\newblock In {\em ECCV}. 2004.

\bibitem{girshick2014rich}
R.~Girshick, J.~Donahue, T.~Darrell, and J.~Malik.
\newblock Rich feature hierarchies for accurate object detection and semantic
  segmentation.
\newblock In {\em CVPR}, 2014.

\bibitem{glorot2010understanding}
X.~Glorot and Y.~Bengio.
\newblock Understanding the difficulty of training deep feedforward neural
  networks.
\newblock In {\em AISTATS}, 2010.

\bibitem{golge2014conceptmap}
E.~Golge and P.~Duygulu.
\newblock Conceptmap: Mining noisy web data for concept learning.
\newblock In {\em ECCV}. 2014.

\bibitem{hariharan2014simultaneous}
B.~Hariharan, P.~Arbel{\'a}ez, R.~Girshick, and J.~Malik.
\newblock Simultaneous detection and segmentation.
\newblock In {\em ECCV}. 2014.

\bibitem{hariharan2012discriminative}
B.~Hariharan, J.~Malik, and D.~Ramanan.
\newblock Discriminative decorrelation for clustering and classification.
\newblock In {\em ECCV}.

\bibitem{hoiem2012diagnosing}
D.~Hoiem, Y.~Chodpathumwan, and Q.~Dai.
\newblock Diagnosing error in object detectors.
\newblock In {\em ECCV}. 2012.

\bibitem{jia2014caffe}
Y.~Jia, E.~Shelhamer, J.~Donahue, S.~Karayev, J.~Long, R.~Girshick,
  S.~Guadarrama, and T.~Darrell.
\newblock Caffe: Convolutional architecture for fast feature embedding.
\newblock In {\em ACM MM}, 2014.

\bibitem{krizhevsky2012imagenet}
A.~Krizhevsky, I.~Sutskever, and G.~E. Hinton.
\newblock Imagenet classification with deep convolutional neural networks.
\newblock In {\em NIPS}, 2012.

\bibitem{kumar2010self}
M.~P. Kumar, B.~Packer, and D.~Koller.
\newblock Self-paced learning for latent variable models.
\newblock In {\em NIPS}, 2010.

\bibitem{lee2011learning}
Y.~J. Lee and K.~Grauman.
\newblock Learning the easy things first: Self-paced visual category discovery.
\newblock In {\em CVPR}, 2011.

\bibitem{li2010optimol}
L.-J. Li and L.~Fei-Fei.
\newblock {OPTIMOL}: automatic online picture collection via incremental model
  learning.
\newblock {\em IJCV}, 2010.

\bibitem{li2013harvesting}
Q.~Li, J.~Wu, and Z.~Tu.
\newblock Harvesting mid-level visual concepts from large-scale internet
  images.
\newblock In {\em CVPR}, 2013.

\bibitem{lin2014microsoft}
T.-Y. Lin, M.~Maire, S.~Belongie, J.~Hays, P.~Perona, D.~Ramanan,
  P.~Doll{\'a}r, and C.~L. Zitnick.
\newblock Microsoft coco: Common objects in context.
\newblock In {\em ECCV}. 2014.

\bibitem{mezuman2012learning}
E.~Mezuman and Y.~Weiss.
\newblock Learning about canonical views from internet image collections.
\newblock In {\em NIPS}, 2012.

\bibitem{miller1995wordnet}
G.~A. Miller.
\newblock Wordnet: a lexical database for english.
\newblock {\em Communications of the ACM}, 1995.

\bibitem{Oquab13}
M.~Oquab, L.~Bottou, I.~Laptev, and J.~Sivic.
\newblock Weakly supervised object recognition with convolutional neural
  networks.
\newblock Technical report, 2014.

\bibitem{ordonez2011im2text}
V.~Ordonez, G.~Kulkarni, and T.~L. Berg.
\newblock Im2text: Describing images using 1 million captioned photographs.
\newblock In {\em NIPS}, 2011.

\bibitem{pandey2011scene}
M.~Pandey and S.~Lazebnik.
\newblock Scene recognition and weakly supervised object localization with
  deformable part-based models.
\newblock In {\em ICCV}, 2011.

\bibitem{papandreou2015weakly}
G.~Papandreou, L.-C. Chen, K.~Murphy, and A.~L. Yuille.
\newblock Weakly-and semi-supervised learning of a dcnn for semantic image
  segmentation.
\newblock {\em arXiv:1502.02734}, 2015.

\bibitem{pathak2014fully}
D.~Pathak, E.~Shelhamer, J.~Long, and T.~Darrell.
\newblock Fully convolutional multi-class multiple instance learning.
\newblock {\em arXiv:1412.7144}, 2014.

\bibitem{quattoni2009recognizing}
A.~Quattoni and A.~Torralba.
\newblock Recognizing indoor scenes.
\newblock In {\em CVPR}, 2009.

\bibitem{raguram2008computing}
R.~Raguram and S.~Lazebnik.
\newblock Computing iconic summaries of general visual concepts.
\newblock In {\em CVPRW}, 2008.

\bibitem{razavian2014cnn}
A.~S. Razavian, H.~Azizpour, J.~Sullivan, and S.~Carlsson.
\newblock Cnn features off-the-shelf: an astounding baseline for recognition.
\newblock In {\em CVPRW}, 2014.

\bibitem{reed2014training}
S.~Reed, H.~Lee, D.~Anguelov, C.~Szegedy, D.~Erhan, and A.~Rabinovich.
\newblock Training deep neural networks on noisy labels with bootstrapping.
\newblock {\em arXiv:1412.6596}, 2014.

\bibitem{russakovsky2014imagenet}
O.~Russakovsky, J.~Deng, H.~Su, J.~Krause, S.~Satheesh, S.~Ma, Z.~Huang,
  A.~Karpathy, A.~Khosla, M.~Bernstein, et~al.
\newblock Imagenet large scale visual recognition challenge.
\newblock {\em arXiv:1409.0575}, 2014.

\bibitem{saenko2009unsupervised}
K.~Saenko and T.~Darrell.
\newblock Unsupervised learning of visual sense models for polysemous words.
\newblock In {\em NIPS}, 2009.

\bibitem{schroff2011harvesting}
F.~Schroff, A.~Criminisi, and A.~Zisserman.
\newblock Harvesting image databases from the web.
\newblock {\em TPAMI}, 2011.

\bibitem{simonyan2013deep}
K.~Simonyan, A.~Vedaldi, and A.~Zisserman.
\newblock Deep inside convolutional networks: Visualising image classification
  models and saliency maps.
\newblock {\em arXiv:1312.6034}, 2013.

\bibitem{simonyan2014very}
K.~Simonyan and A.~Zisserman.
\newblock Very deep convolutional networks for large-scale image recognition.
\newblock {\em arXiv:1409.1556}, 2014.

\bibitem{sivic2003video}
J.~Sivic and A.~Zisserman.
\newblock Video google: A text retrieval approach to object matching in videos.
\newblock In {\em ICCV}, 2003.

\bibitem{smeulders2000content}
A.~W. Smeulders, M.~Worring, S.~Santini, A.~Gupta, and R.~Jain.
\newblock Content-based image retrieval at the end of the early years.
\newblock {\em TPAMI}, 2000.

\bibitem{song2014learning}
H.~O. Song, R.~Girshick, S.~Jegelka, J.~Mairal, Z.~Harchaoui, and T.~Darrell.
\newblock On learning to localize objects with minimal supervision.
\newblock In {\em ICML}.

\bibitem{sukhbaatar2014learning}
S.~Sukhbaatar and R.~Fergus.
\newblock Learning from noisy labels with deep neural networks.
\newblock {\em arXiv:1406.2080}, 2014.

\bibitem{szegedy2014going}
C.~Szegedy, W.~Liu, Y.~Jia, P.~Sermanet, S.~Reed, D.~Anguelov, D.~Erhan,
  V.~Vanhoucke, and A.~Rabinovich.
\newblock Going deeper with convolutions.
\newblock {\em arXiv:1409.4842}, 2014.

\bibitem{taigman2014deepface}
Y.~Taigman, M.~Yang, M.~Ranzato, and L.~Wolf.
\newblock Deepface: Closing the gap to human-level performance in face
  verification.
\newblock In {\em CVPR}, 2014.

\bibitem{torralba2011unbiased}
A.~Torralba and A.~A. Efros.
\newblock Unbiased look at dataset bias.
\newblock In {\em CVPR}, 2011.

\bibitem{torresani2010efficient}
L.~Torresani, M.~Szummer, and A.~Fitzgibbon.
\newblock Efficient object category recognition using classemes.
\newblock In {\em ECCV}. 2010.

\bibitem{vijayanarasimhan2008keywords}
S.~Vijayanarasimhan and K.~Grauman.
\newblock Keywords to visual categories: Multiple-instance learning forweakly
  supervised object categorization.
\newblock In {\em CVPR}, 2008.

\bibitem{wang2014weakly}
C.~Wang, W.~Ren, K.~Huang, and T.~Tan.
\newblock Weakly supervised object localization with latent category learning.
\newblock In {\em ECCV}. 2014.

\bibitem{wang2008annotating}
X.-J. Wang, L.~Zhang, X.~Li, and W.-Y. Ma.
\newblock Annotating images by mining image search results.
\newblock {\em TPAMI}, 2008.

\bibitem{xia2014well}
Y.~Xia, X.~Cao, F.~Wen, and J.~Sun.
\newblock Well begun is half done: Generating high-quality seeds for automatic
  image dataset construction from web.
\newblock In {\em ECCV}. 2014.

\bibitem{xiao2010sun}
J.~Xiao, J.~Hays, K.~A. Ehinger, A.~Oliva, and A.~Torralba.
\newblock Sun database: Large-scale scene recognition from abbey to zoo.
\newblock In {\em CVPR}, 2010.

\bibitem{zhou2014learning}
B.~Zhou, A.~Lapedriza, J.~Xiao, A.~Torralba, and A.~Oliva.
\newblock Learning deep features for scene recognition using places database.
\newblock In {\em NIPS}, 2014.

\bibitem{zitnick2014edge}
C.~L. Zitnick and P.~Doll{\'a}r.
\newblock Edge boxes: Locating object proposals from edges.
\newblock In {\em ECCV}. 2014.

\end{thebibliography}
% }

\bibliographystyle{ieee}

\end{document}